\documentclass[parskip=full,fleqn,11pt]{article}

\usepackage{amssymb}
\usepackage{amsthm}
\usepackage{amsmath}
\usepackage{float,booktabs}
\usepackage{float} 
\usepackage{graphicx}
\usepackage{natbib}
\usepackage{url}
\usepackage{xcolor}
\usepackage{soul,color,colortbl}
\usepackage{array,multirow}
\usepackage{cancel}
\usepackage[top=0.9in, bottom=1.2in, left=0.875in, right=0.875in]{geometry}
\usepackage[linesnumbered,ruled]{algorithm2e}
\usepackage{tikz}
\usetikzlibrary{shapes.geometric}
\usetikzlibrary{arrows.meta,arrows}
\usetikzlibrary{er,positioning,arrows.meta,calc}
\usepackage[labelformat=simple]{subcaption}
\usepackage{bm}
\usepackage[linesnumbered,ruled]{algorithm2e}
\usepackage{caption}
\usepackage{subcaption}
\usepackage{bm}
\usepackage{bbm}

\newcommand{\EE}{\text{E}}

\newtheorem{theorem}{Theorem}[section]

\newtheorem{lemma}[theorem]{Lemma}

\newtheorem{proposition}[theorem]{Proposition}

\def\spacingset#1{\renewcommand{\baselinestretch}
{#1}\small\normalsize} \spacingset{1}

\newcolumntype{P}[1]{>{\centering\arraybackslash}p{#1}}
\newcommand*{\myfont}{\fontfamily{lmss}\selectfont}
\DeclareTextFontCommand{\textpython}{\myfont}
\DeclareMathOperator*{\argmin}{\arg\!\min}
\DeclareMathOperator*{\argmax}{\arg\!\max}

\DeclareCaptionLabelFormat{boldnumber}{\bothIfFirst{#1}{~}\textbf{#2}}
\title{\textbf{The SAMME.C2 algorithm for severely imbalanced multi-class classification}}
\author{Banghee So \thanks{Department of Mathematics, Towson University, 7800 York Rd, Towson, MD, 21252, USA. Email: \texttt{bso@towson.edu}.} 
\and Emiliano A. Valdez\thanks{Corresponding author; Department of Mathematics, University of Connecticut, 341 Mansfield Road, Storrs, CT, 06269-1009, USA. Email: \texttt{emiliano.valdez@uconn.edu}.}}

\begin{document}

\maketitle

\begin{abstract}

Classification predictive modeling involves the accurate assignment of observations in a dataset to target classes or categories. There is an increasing growth of real-world classification problems with severely imbalanced class distributions. In this case, minority classes have much fewer observations to learn from than those from majority classes. Despite this sparsity, a minority class is often considered the more interesting class yet developing a scientific learning algorithm suitable for the observations presents countless challenges. In this article, we suggest a novel multi-class classification algorithm specialized to handle severely imbalanced classes based on the method we refer to as \texttt{SAMME.C2}. It blends the flexible mechanics of the boosting techniques from \texttt{SAMME} algorithm, a multi-class classifier, and \texttt{Ada.C2} algorithm, a cost-sensitive binary classifier designed to address highly class imbalances. Not only do we provide the resulting algorithm but we also establish scientific and statistical formulation of our proposed \texttt{SAMME.C2} algorithm. Through numerical experiments examining various degrees of classifier difficulty, we demonstrate consistent superior performance of our proposed model.

\vspace{1.0cm}

\noindent \textbf{Keywords}: AdaBoost (Adaptive boosting); Cost-sensitive learning; Forward stagewise additive modeling; \texttt{SAMME} (Stagewise Additive Modeling using a Multi-class Exponential Loss Function); \texttt{SAMME.C2}; \texttt{SMOTE} (Synthetic Minority Oversampling Technique)

\end{abstract}

\newpage

\section{Introduction} \label{sec:intro}

In machine learning, classification involves the accurate assignment of a target class or label to input observations. When there are only two labels, it is called binary classification; when there are more than two labels, it is often referred to as multi-class classification. Classification algorithms may generally fall into three categories (see \citet{hastie2009}):
\begin{itemize}
    \item Linear classifiers: This type of algorithm separates the input observations using a line or hyperplane based on a linear combination of the input features. Examples in this algorithm include logistic regression, probit regression, and linear discriminant analysis.
    \item Nonlinear classifiers: This type of algorithm separates the input observations based on nonlinear functions. Examples include decision trees, $k$-nearest neighbors (KNN), support vector machines (SVM), and neural networks.
    \item Ensemble methods: This type of algorithm combines the predictions produced from multiple models. Examples includes random forests, stochastic gradient boosting (e.g., XGBoost), and adaptive boosting (AdaBoost).
\end{itemize}

In imbalanced classification, the distribution of observations across classes is biased or skewed. In this case, minority classes have much fewer observations to learn from than those from majority classes. In spite of the sparsity, the minority class is often considered the more interesting class yet developing a scientific learning algorithm suitable for the observations presents countless challenges. Several research have been conducted dealing with imbalanced data, yet mostly in the context of binary problems. The commonplace and more direct approach is to use algorithms listed above and handle the class imbalance at the data level. In this case, the class distribution of the input observations is rebalanced by oversampling (or undersampling) from the underrepresented (or overrepresented) classes. One popular approach is the oversampling of underrepresented classes based on the \texttt{SMOTE} (Synthetic Minority Oversampling Technique), a technique developed by \citet{chawla2002smote}. It is worth noting that generating synthetic observations to rebalance class distributions, especially with multi-class classification, has the disadvantage of increasing the overlapping classes with unnecessary additional noise.

One popular class of algorithms that is believed to be one of the most powerful techniques is boosting, which is based on training a sequence of weak models into a strong learner in order to improve predictive power. A specific boosting technique primarily developed for classification is AdaBoost, a class of so-called adaptive boosting algorithms. \texttt{AdaBoost.M1}, which combines several weak classifiers to produce a strong classifier, is the first practical boosting algorithm introduced by \citet{freund1997decision}. \texttt{AdaBoost.M1} is an iterative process that starts with a distribution of equal observation weights. At each iteration, the process fits one weak classifier and subsequently adjusts the observation weights based on the idea that more weights are given to input observations that have been misclassified, allowing for increased learning. See Algorithm \ref{alg:ada.m1} in Appendix A.

\texttt{AdaBoost.M1} has been extended to handle multi-class classification problems. One such extension is the so-called \texttt{AdaBoost.M2}, developed by \citet{freund1997decision} that is based on the optimization of a pseudo-loss function suitable for handling multi-class problems. Another extension is the \texttt{AdaBoost.MH} developed by \citet{schapire1999improv} that is based on the optimization of the Hamming loss function. Both these extensions solve multi-class classification problems by reducing them into several different binary problems; Such procedures can be slow and inefficient. A more popular multi-class AdaBoost extension is the algorithm called \texttt{SAMME} (Stagewise Additive Modeling using a Multi-class Exponential Loss Function) proposed by \citet{zhu2009mclass}, which avoids computational inefficiencies without the multiple binary problems. See \citet{so2021cost} for details of the iteration process of this algorithm. According to \citet{friedman2000addlog} and \citet{hastie2009}, the \texttt{SAMME} algorithm is equivalent to an additive model with a minimization of a multi-class exponential loss function and belongs to the traditional statistical family of forward stagewise additive models. Additional variations to these AdaBoost algorithms have appeared in \citet{ferreira2012review}; a recent work of \citet{tanha2020boosting} provides a comprehensive survey.

In order to further improve prediction within an imbalanced classification, cost-sensitive learning algorithms provides for a necessary additional layer of complexity in the algorithm that takes costs into consideration. The work of \citet{pazzani1994reduce} was the first to introduce cost-sensitive algorithms that minimize misclassification costs in classification problems. The cost values, estimated as hyperparameters, are additional inputs to the learning procedure and are generally used to reduce misclassification costs, which attach penalty to predictions that lead to significant errors. These costs indeed are used to modify the updating of the observation weights at each iteration within the context of adaptive boosting algorithms. For binary classification, \texttt{Ada.C2} is the most well-known and attractive method of AdaBoost that combines cost-sensitive learning (\citet{sun2007cost}). For details of this algorithm, please see \citet{so2021cost}. 

In this article, we suggest a novel multi-class classification algorithm, which we refer to as \texttt{SAMME.C2}, especially designed to handle imbalanced classes. This algorithm is inspired by combining the advantages drawn from two algorithms we earlier described: (1) \texttt{SAMME}, one of the Adaboost algorithms for multi-class classifiers that do not decompose the classification task into multiple binary classes to avoid computational inefficiencies, and (2) cost-sensitive learning employed in \texttt{Ada.C2}. \citet{zhu2009mclass} showed that \texttt{SAMME} is equivalent to a forward stagewise additive modeling with a minimization of a multi-class exponential loss function and has been proven to be Bayes classifier. These mathematical proofs are important statistical justifications that the resulting classifiers are optimal. However, we find that the training purpose of the \texttt{SAMME} algorithm is to reduce test error rates and this works quite well when classes are generally considered balanced. In the case when classes are severely imbalanced, the \texttt{SAMME} algorithm places more observation weight on classifying majority classes accurately because this contributes more on decreasing test errors. Further, this results in a huge sacrifice of being able to accurately classify minority classes. This leads us to embrace the idea of adding the attributes of cost-sensitive learning techniques to this algorithm. When cost-sensitive learning is added to \texttt{SAMME},  \texttt{SAMME.C2} is able to demonstrate the superiority of controlling these peculiar issues attributable to class imbalances. This article extends the mathematical proofs that with the addition of cost values, \texttt{SAMME.C2} retains the same statistical foundations with \texttt{SAMME}.

The practical importance of multi-class classification tasks, especially with severely imbalanced classes, extends to multiple disciplines. Various ad-hoc algorithms, some of which are described above, have been employed. The works of \citet{liu2017hybrid}), \citet{yuan2018regularized}, \citet{jeong2020comparison}, and \citet{mahmudah2021machine} address real life biomedical applications of such classification tasks in the detection of disease. Spam detection is widely studied in computer engineering; see \citet{mohammad2020improved},  \citet{talpur2020multi}, and \citet{dewi2017multiclass}. The research probe conducted by \citet{kim2016detecting} applies multi-class classification tasks with cost-sensitive learning mechanisms to detect financial misstatements associated with fraud intention in finance. In operations research, \citet{han2019fault} proposes a fault diagnosis model for planetary gear carrier packs as a detection tool for manufacture fault. Finally, in insurance, \citet{so2021cost} examines the frequency of accidents as a multi-class classification problem with a highly imbalanced class using observations of insured drivers with additional telematics information about driving behavior through a usage-based insurance policy.

The remainder of the papers is as follows. Section \ref{sec:sc2} introduces the details about this new \texttt{SAMME.C2} algorithm, which is largely based on the integration of \texttt{SAMME} and \texttt{Ada.C2}. Section \ref{sec:theory} presents the mathematical proofs that \texttt{SAMME.C2} follows a forward stagewise additive model and is an optimal Bayes classifier. To demonstrate the algorithmic superiority of \texttt{SAMME.C2}, Section \ref{sec:exp} presents numerical experiment results based on simulated datasets. To show the many varied applications of our work, this section additionally lists some practical researches on multi-class classification. Section \ref{sec:conclude} concludes the chapter. 

\section{The \texttt{SAMME.C2} algorithm} \label{sec:sc2}

For our purpose, let us consider a set of $N$ input observations denoted by $(\bm{x}_i,y_i)$ for $i=1,\ldots,N$ where $\bm{x}_i$ is a set of feature variables and $y_i \in Y=\{1,2,\ldots,K\}$ is target classification variable belonging to one of $K$ classes. In the case of binary, $K=2$. An important input variable is the cost value for which we denote here as $C(y_i)$ to emphasize that it is a function of the target variable pre-determined by hyperparameter optimization technique described below.

\texttt{SAMME.C2} combines the benefits of boosting and cost-sensitive algorithms for handling class imbalances in multi-class classification problems. Given the input data $(\bm{x}_i,y_i,C(y_i))$, the algorithm is an iterative process of fitting weak classifiers denoted by $h_t(\bm{x}_i)$ at iteration $t$ and the process stops at time $T$. 
The stopping time $T$ can be a tuned hyperparameter. At iteration $t=1$, we set equal observation weights as $D_1(i)=\frac{1}{N}$. In subsequent iteration $t$, we train weak classifiers using the distribution $D_t$. Any weak classifier can be used but for our purpose, the simplest weak classifiers are decision stumps. We update the distribution of the observation weights using
\begin{equation} \label{eq:1}
D_{t+1}(i) = \dfrac{C(y_i) \, D_t(i) \exp(-\alpha_t I(y_i = h_t(\bm{x}_i)))}{\sum_{j=1}^{N} C(y_j) \, D_t(j) \exp(-\alpha_t I(y_j = h_t(\bm{x}_j)))}
\end{equation}
which depends on the error rate of the $t$-th weak classifier given by
\begin{equation} \label{eq:2}
	\epsilon_t = \frac{\sum_{i=1}^{N} D_t(i) I(y_i \ne h_t(\bm{x}_i))}{\sum_{i=1}^m D_t(i)},
\end{equation}
and the weight of the $t$-th weak classifier given by
\begin{equation} \label{eq:3}
	\alpha_t = \log\!\Big(\dfrac{1-\epsilon_t}{\epsilon_t}\Big) + \log(K-1).
\end{equation}
The final classifier is then determined at the final iteration $T$ using
\begin{equation} \label{eq:4}
H(\bm{x}_i)= \underset{k}{\mathrm{argmax}} {\  \sum_{t=1}^{T} \alpha_t I(h_t(\bm{x}_i) = k)}
\end{equation}
For details of the algorithm is in Algorithm \ref{alg:sammec2} in the appendix.

\subsection{Comparison with \texttt{Ada.C2} and \texttt{SAMME}} \label{sub.compare}

The iteration process for all the three algorithms (\texttt{Ada.C2}, \texttt{SAMME}, and \texttt{SAMME.C2}) are exactly the same. However, the primary differences lie in the comparison of the error rate and the weight of the $t$-th classifier, as well as the updating of the distribution of the observation weights.

For \texttt{SAMME} and \texttt{SAMME.C2}, the computation of the error rate is exactly the same despite that the \texttt{SAMME} algorithm does not have cost values. For the \texttt{Ada.C2}, unlike \texttt{SAMME.C2}, the cost values are used to compute error rate of the $t$-th classifier using
\begin{equation} \label{eq:5}
	\epsilon_t = \frac{\sum_{i=1}^{N} C(y_i) D_t(i) I(y_i \ne h_t(\bm{x}_i))}{\sum_{i=1}^m C(y_i) D_t(i)}.
\end{equation}

For \texttt{SAMME} and \texttt{SAMME.C2}, the computation of the weight of the $t$-th classifier is exactly the same despite that the \texttt{SAMME} algorithm does not have cost values. For the \texttt{Ada.C2}, the weight of the $t$-th classifier is given by
\begin{equation} \label{eq:6}
	\alpha_t = \dfrac{1}{2}\log\!\Big(\dfrac{1-\epsilon_t}{\epsilon_t}\Big).
\end{equation}
For the misclassified training samples to be properly boosted, the classification error at each iteration should be less than 1/2, otherwise, $\alpha_t$, which is a function of the classification error will be negative and observation weights will be updated in the wrong direction. In which case, after the iteration, the classification error can no longer be improved. In the case of binary as in \texttt{Ada.C2}, this just requires that each weak learner performs a little better than random guessing. However, when $K>2$, the random guessing accuracy rate is $1/K$, which is less than 1/2. Hence, multi-class problems need much more accurate weak learner than binary problem, and if weak learner is not chosen and trained accurately enough, the algorithm may fail. \citet{zhu2009mclass} pointed this out and suggested \texttt{SAMME} algorithm, which directly extend \texttt{AdaBoost.M1} to the multi-class cases as adding one term, $\log(K-1)$, to the updating equation of $\alpha_t$ at each iteration $t$. 

The updating of the distribution of the observation weights for the subsequent iteration is exactly the same for the \texttt{Ada.C2} and \texttt{SAMME.C2} algorithms; this is not at all surprising since both algorithms consider cost values. For the \texttt{SAMME} algorithm for which it does not have cost values, the distribution of the observation weights for the subsequent iteration is given by
\begin{equation} \label{eq:7}
D_{t+1}(i) = \dfrac{D_t(i) \exp(-\alpha_t I(y_i = h_t(\bm{x}_i)))}{\sum_{j=1}^{N} D_t(j) \exp(-\alpha_t I(y_j = h_t(\bm{x}_j)))}.
\end{equation}
The updating principle is based on the idea on how the algorithm correctly classifies (or misclassifies) majority and minority classes. For the \texttt{SAMME} algorithm without cost values, there is an even redistribution of correct classification (or misclassification) regardless of whether it belongs to a majority or minority class. For the \texttt{Ada.C2} and \texttt{SAMME.C2}, with addition of cost values, the redistribution becomes uneven by assigning heavier weights to observations that belong to minority classes. This leads us to conclude that after enough number of iterations, for cost-sensitive learning mechanisms, weak classifiers are trained with a heavy emphasis on misclassified observations that are in the minority class. See Figure 1 of \citet{so2021cost}.

For a graphical display of the iteration process with emphasis on these differences, please refer to Figure \ref{fig:three algorithms}. It can be noted that \texttt{SAMME} is a special case of \texttt{SAMME.C2} by assigning all the cost values to 1, that is, $C(y_i)=1$ for all $y_i \in Y = \{1,2,\ldots,K\}$ and $i=1,2,\ldots,N$.

	\tikzstyle{block} = [rectangle, draw, fill=blue!20, 
	text width=32em, text centered, minimum height=4em]
	\tikzstyle{line} = [draw, -latex']

	\begin{center}
			\resizebox{!}{11.5cm}{
			\begin{tikzpicture}[node distance = 4.5cm, auto]
				\node [block] (init) {Initial sample weights $D_1(i) = \frac{1}{N}, \quad i =1,\ldots,N$};
				\node [block, right of=init, node distance=14cm] (weak) {Train weak classifier using the distribution $D_t$, for $t=1,\ldots,T$};
				\node [block, below of=init] (dist) { Get weak classifier $h_t\!:$ $X \rightarrow k \in \{1,\ldots,K\}$ };
				\node [block, below of=dist] (train) {Compute error rate of $h_t$
					$$\texttt{\color{blue}SAMME \color{black}\& \color{purple}SAMME.C2: }\color{purple}\epsilon_t = \frac{\sum_{i=1}^{N} D_t(i) I(y_i \ne h_t(\bm{x}_i))}{\sum_{i=1}^N D_t(i)}$$
					$$\color{blue}\texttt{Ada.C2: }\epsilon_t = \frac{\sum_{i=1}^{N} C(y_i) \, D_t(i) I(y_i \ne h_t(\bm{x}_i))}{\sum_{i=1}^N C(y_i) \, D_t(i)}$$};
				\node [block, below of=train,node distance=6.5cm] (calc) {Calculate weight of the $t$-th weak classifier
					$$\texttt{\color{blue}SAMME \color{black}\& \color{purple}SAMME.C2: }\color{purple}\alpha_t = \log\!\Big(\frac{1-\epsilon_t}{\epsilon_t}\Big) + \log(K-1)$$ 
					$$\color{blue}\texttt{Ada.C2: }\alpha_t = \frac{1}{2} \log\!\Big(\frac{1-\epsilon_t}{\epsilon_t}\Big)$$};
				\node [block, right of=train, node distance=14cm] (update) {Update sample weights
					$$\texttt{\color{blue}Ada.C2 \color{black}\& \color{purple}SAMME.C2: } \color{purple}D_{t+1}(i) = \dfrac{C(y_i) \, D_t(i) \exp(-\alpha_t I(y_i = h_t(\bm{x}_i)))}{\sum_{j=1}^{N} C(y_j) \, D_t(j) \exp(-\alpha_t I(y_j = h_t(\bm{x}_j)))}$$
					$$\color{blue}\texttt{SAMME: }D_{t+1}(i) = \dfrac{D_t(i) \exp(-\alpha_t I(y_i = h_t(\bm{x}_i)))}{\sum_{j=1}^{N} D_t(j) \exp(-\alpha_t I(y_j = h_t(\bm{x}_j)))}$$};
				\node [block, below of=update,node distance=6.5cm] (return) {Return (final) classifier
					$$H(\bm{x}_i) = \underset{k}{\mathrm{argmax}} {\  \sum_{t=1}^{T} \alpha_t I(h_t(\bm{x}_i) = k)}$$};
				\path [line] (init) -- (weak);
				\path [line, line width=1.5pt] (weak) -- (dist);
				\path [line, line width=1.5pt] (dist) -- (train);
				\path [line, line width=1.5pt] (train) -- (calc);
				\path [line, line width=1.5pt] (calc) -- (update);
				\path [line] (calc) -- (return);
				\path [line, line width=1.5pt] (update) -- (weak);
		\end{tikzpicture}}
	\captionof{figure}{\textbf{Three AdaBoost algorithms: \texttt{SAMME}, \texttt{Ada.C2}, and \texttt{SAMME.C2}}}\label{fig:three algorithms}
	\end{center}

\subsection{The cost optimization} \label{sub.cost}

The critical work involved in implementing \texttt{SAMME.C2} is the process of determining the cost value given to each class. From the perspective of \texttt{SAMME.C2}, because cost values can be regarded as hyperparameters, this process can be regarded as optimizing or tuning a hyperparameter in a learning algorithm. Various optimization methods of hyperparameters may be used to optimize the cost values. Some of the frequently used optimizing strategies are grid search, random search (\citet{bergstra2012randomsearch}), and sequential model-based optimization (\citet{bergstra2011smbo}). The simple and widely used optimization algorithms are the grid search and the random search.  However, since the next trial set of hyperparameters are not chosen based on previous results, it is time-consuming. One of the most powerful strategies is the sequential model-based optimization, also sometimes referred to as Bayesian optimization. The subsequent set of hyperparameters is determined based on the result of the previously determined sets of hyperparameters.

\citet{bergstra2011smbo} and \citet{snoek2012practical} showed that sequential model-based optimization outperforms both grid and random searches. However, to use the sequential model-based optimization, advanced level of statistical knowledge is required. For our purpose with \texttt{SAMME.C2}, we employ Genetic Algorithm (GA), which is simple, easily understandable, and at the same time, computationally efficient.  Developed by \citet{holland1975} and described in \citet{muhlenbein1997GA}, GA is one kind of random search techniques, but the primary difference from general random searches is that the subsequent trial set of hyperparameters are decided based on the result of previously determined sets of hyperparameters just like the sequential model-based optimization.

In this algorithm, we first create the population set consisting of $M$ arbitrary cost vectors. The cost vector has $K$ elements for $K$-class problem. We then run \texttt{SAMME.C2} and perform evaluation step to get the performance metric corresponding to each cost vector. Here, performance metric is referred to as the objective function.
\begin{itemize}
\item In the selection step, two cost vectors are chosen from the $M$ vectors, employing the ``choice by roulette'' method typically used as an operator in GA algorithm with the objective of selecting cost vectors having a larger performance metric with a higher possibility.
\item In the crossover step, we combine the selected two cost vectors into a single vector using arithmetic average.
\item In the mutation step, we pick a random number within a tiny interval that is used to adjust the elements in the cost vector. 
\end{itemize}
Repeating this selection, crossover, and mutation steps, we can produce a new population with new $M$ cost vectors, for which the procedure is iteratively repeated $P$ number of times to generate the population that will produce the optimal cost vectors.

\section{Proof of optimality} \label{sec:theory}

In this section, we provide a theoretical justification of the \texttt{SAMME.C2} algorithm. Recall that an advantage of the \texttt{SAMME} algorithm is that it is statistically explainable or justifiable. In particular, \citet{zhu2009mclass} proved that the \texttt{SAMME} algorithm is equivalent to fitting a forward stagewise additive model using a multi-class exponential loss function expressed as
\begin{equation}
	L(\bm{U},\bm{f(x)})\;= \; \exp \left(-\frac{1}{K}\bm{U}'\bm{f(x)}\right) \label{eq:8}.
\end{equation}
In the same fashion, we demonstrate that the addition of cost sensitive learning to \texttt{SAMME} preserves these same theoretical aspects. To prove this, instead of (\ref{eq:8}), we use a loss function multiplied by cost values, which we may call a multi-class cost sensitive exponential loss function expressed as
\begin{equation}
	L(\bm{U},\bm{f(x)},\bm{C})\;= \;\bm{C} \exp \left(-\frac{1}{K}\bm{U}'\bm{f(x)}\right). \label{eq:9}
\end{equation}
Just as in the work of \citet{zhu2009mclass}, we justify the use of multi-class cost-sensitive exponential loss function in (\ref{eq:9}) by first showing that the resulting classifier minimizing (\ref{eq:9}) is the optimal Bayes classifier. Note that the symbols $\bm{U}$, $\bm{f(x)}$, and cost vector $\bm{C}$ will be well defined in the subsequent subsections.

\subsection{Terminology} \label{sub.term}

Suppose we are given a set of data denoted by $\mathcal{D}=(\bm{x}_i,y_i,C(y_i))$ for $i=1,2,\ldots,N$ where $\bm{x}_i$ is a set of feature variables,  $y_i$ is the corresponding response which is a classification variable that belongs to the set $Y =\{1, 2,\ldots ,K\}$ and $C(y_i)$ is the corresponding cost value which is a function of $y_i$. For each observation, we attach a cost value that depends on which class observation $i$ belongs to and these are generated outside the algorithm but are based on the minority/majority characteristics of the classification variable. The objective is to learn from the data so that we can build a predictive model for identifying a particular observation will belong to a particular class, given the set of feature variables.
Without loss of generality, we re-code the response $y_i = \bm{U}_i$; all entries in this vector will be equal to $-1/(K-1)$ except a value of 1 in position $k$ if the observation $y_i = k$. In effect, we have $\bm{U}_i = (U_{1i}, U_{2i},\ldots,U_{Ki})'$ where
\[
	U_{ki}=
	\begin{cases}
		1, & \text{if}\ y_i=k \\
		-\frac{1}{K-1}, & \text{if}\ y_i \neq k
	\end{cases}
\]
This re-coding is for carrying over the symmetry of class label representation in the binary case (\citet{lee2004multicategory}). It is straightforward to show that $\sum_{j=1}^{K}U_{ji}=0$ for all $i=1,2,\ldots, N$. There is a one-to-one correspondence between $\mathbf{U}_i$ and $y_i$ and will be interchangeably used for convenience and clarity whenever possible; each equivalently refers to the class the observation $i$ belongs to.

\subsection{The loss function for the optimal Bayes classifier} \label{sub.bayes}

This section provides a theoretical justification for the use of the multi-class cost-sensitive exponential loss function in (\ref{eq:9}) in the optimization leading to the \texttt{SAMME.C2} algorithm. More specifically, we show here that the resulting classifier is an optimal Bayes classifier. It is well-known in classification problems that this produces a classifier that minimizes the probability of misclassification. See \citet{hastie2009}.

\begin{lemma} \label{lem1}
Denote $y$ to be the classification variable with possible values belonging to $\{1,2,\ldots,K\}$, $\bm{U}$ to be the re-coding of this variable as explained above, $\bm{C}=(C_1,C_2,\ldots,C_K)$ to be the cost vector, and \[
\bm{f(x)} = (f_1(\bm{x}), f_2(\bm{x}), \ldots, f_K(\bm{x}))
\]
to be the classifier function. The following result leads us to the optimal classifier function under the multi-class cost-sensitive exponential loss function:
\[
\underset{\bm{f(x)}}{\argmin}\;\;\EE_{\bm{U}|\bm{x}} \! \left[\bm{C} \exp \left(-\frac{1}{K}\bm{U}'\bm{f(x)}\right)\right] = (f^{*}_1(\bm{x}),f^{*}_2(\bm{x}),\ldots,f^{*}_K(\bm{x})),
\]
subject to $f_1+\ldots+f_K=0$ where
\begin{equation}
f^{*}_k(\bm{x})=(K-1)\bigg(\log\text{Prob}(y=k|\bm{x})-\frac{1}{K}\sum_{j=1}^{K}\log\text{Prob}(y=j|\bm{x})\bigg),\quad k=1,2,\ldots,K. \label{eq:10}
\end{equation}
\begin{proof}
For this optimization, the Lagrange can be written as
\[
\sum_{k=1}^{K} \left[C_k\exp\left(-\frac{1}{K}\left(f_k -\frac{\sum_{j\ne k}f_j}{K-1}\right)\right) \text{Prob}(y=k|\bm{x}) -\lambda f_k \right] = \sum_{k=1}^{K} \left[C_k\exp\left(-\frac{f_k}{K-1}\right) \text{Prob}(y=k|\bm{x}) - \lambda f_k \right]
\]
where $\lambda$ is the Lagrange multiplier. By taking derivative with respect to $f_k$ and $\lambda$, we reach
\[
-\frac{1}{K-1} C_k \exp\left(-\frac{f_k(\bm{x})}{K-1}\right) \text{Prob}(y=k|\bm{x})-\lambda=0, \quad k=1,2,\ldots,K
\]
and the constraint that
\[
f_1(\bm{x})+\ldots+f_K(\bm{x})=0.
\]
Next, by summing the first $K$ equations, we get
\[
\log \left(-\frac{K-1}{C_k}\lambda\right)=\frac{1}{K}\sum_{k=1}^{K}\log\text{Prob}(y=k|\bm{x})
\]
and by substituting the last equation, we obtain the following population minimizer, (\ref{eq:10}):
\[
f^{*}_k(\bm{x})=(K-1)\bigg(\log\text{Prob}(y=k|\bm{x})-\frac{1}{K}\sum_{j=1}^{K}\log\text{Prob}(y=j|\bm{x})\bigg),\quad k=1,2,\ldots,K.
\]
\end{proof}
\end{lemma}

Note that the constraints on $f_k$ in Lemma \ref{lem1} allow us to find the unique solution. The following proposition allows us to choose the optimal Bayes classifier.

\begin{proposition} \label{prop1}
Denote $y$ to be the classification variable with possible values belonging to $\{1,2,\ldots,K\}$. Given the feature variables $\bm{x}$, we find the optimal Bayes classifier using the multi-class cost-sensitive exponential loss function:
\[
y^{*}(\bm{x})=\underset{k}{\argmax} \; \text{Prob}(y=k|\bm{x}).
\]
\begin{proof}
It is clear that
\[
y^{*}(\bm{x})=\underset{k}{\argmax} f^{*}_k(\bm{x})
\]
and that
\[
\frac{1}{K}\sum_{j=1}^{K}\log\text{Prob}(y=j|\bm{x})
\]
is fixed for all $k \in \{1,2,\ldots,K\}$. It follows therefore that from (\ref{eq:10}), we have
\[
\underset{k}{\argmax} \; f^{*}_k(\bm{x}) = \underset{k}{\argmax} \; \text{Prob}(y=k|\bm{x}).
\]
\end{proof}
\end{proposition}

Proposition \ref{prop1} provides a theoretical justification for our estimated classifier in the \texttt{SAMME.C2} algorithm, and the subsequent proposition provides for a formula to calculate the implied class probabilities within this framework.

\begin{proposition}
The implied class probability under the optimal Bayes classifier has the form
\[
\text{Prob}^{*}(y=k|\bm{x})=\displaystyle\frac{\exp(\frac{1}{K-1}f^*_k(\bm{x}))}{\sum_{j=1}^K \exp(\frac{1}{K-1}f^*_j(\bm{x}))}, \quad k=1,2,\ldots,K
\]
\begin{proof}
Equation (\ref{eq:10}) provides $\text{Prob}^{*}(y=k|\bm{x})$ once $f^{*}_k(\bm{x})$'s are determined.
\end{proof}
\end{proposition}

\subsection{\texttt{SAMME.C2} as forward stagewise additive modeling} \label{sub.forward}

In this section, we show that our \texttt{SAMME.C2} algorithm is indeed equivalent to a forward stagewise additive modeling based on the optimization of the multi-class cost-sensitive exponential loss function expressed in (\ref{eq:9}).

Given the training set $\mathcal{D}$, (\ref{eq:9}) can be written as:
\begin{eqnarray*}
	L(\bm{C},\bm{U},\bm{f}(\bm{x})) &=& \bm{C} \exp \left(- \frac{1}{K}\bm{U}'\bm{f}(\bm{x}) \right) \\
	&=& \sum_{i=1}^N C(y_i) \exp \left(-\frac{1}{K}(U_{1i}f_1(\bm{x}_i)+\ldots + U_{Ki}f_K(\bm{x}_i)) \right) \\
	&=& \sum_{i=1}^N C(y_i)  \exp \left(-\frac{1}{K} \mathbf{U}'_i \bm{f}(\bm{x}_i) \right), \\
\end{eqnarray*}
we wish to find $\bm{f}(\bm{x_i})$ such that
\begin{eqnarray}
\argmin_{\bm{f}}  \sum_{i=1}^N C(y_i)  \exp \left(-\frac{1}{K} \mathbf{U}'_i \bm{f}(\bm{x}_i) \right) \label{eq:11}
\end{eqnarray}
subject to $f_1+\ldots+f_K=0$.

Using forward stagewise modeling for learning, the solution to (\ref{eq:11}) has the linear form
\[
\bm{f}(\bm{x_i}) = \sum_{t=1}^{T} \beta^{(t)} \bm{g}^{(t)}(\bm{x_i}),
\]
where $T$ is the total number of iterations, and $\bm{g}^{(t)}(\bm{x_i})$ are the basis functions with corresponding coefficient $\beta^{(t)}$. We require that each basis function satisfies the symmetric constraint whereby $\sum_{k=1}^{K} g^{(t)}_k(\bm{x_i})=0$ for all $t=1,2\ldots,T$ so that $\bm{g}^{(t)}(\bm{x_i})$ takes only one of the possible values from the set
\begin{equation}
	\mathcal{U} = \left\{ \begin{array}{l}
		(1,-\frac{1}{K-1}, \ldots, -\frac{1}{K-1}), \\
		(-\frac{1}{K-1},1, \ldots, -\frac{1}{K-1}), \\
		\vdots \\
		(-\frac{1}{K-1}, \ldots, -\frac{1}{K-1},1)
	\end{array} \right\} \label{eq:12}
\end{equation}
Then, at iteration $t$, the solution can be written as:
\begin{equation}
	\bm{f}^{(t)}(\bm{x}_i) =\underbrace{ \bm{f}^{(t-1)}(\bm{x}_i)}_{=\bm{f}^{(t-2)}(\bm{x}_i)+\beta^{(t-1)} \bm{g}^{(t-1)}(\bm{x}_i)} + \beta^{(t)} \bm{g}^{(t)}(\bm{x}_i), \hspace{0.4 in} \bm{f}^{(0)}(\bm{x_i})=\bm{0}. \label{eq:13}
\end{equation}

Forward stagewise modeling finds the solution to (\ref{eq:11}) by sequentially adding new basis functions to previously fitted model. Hence, at iteration $t$, we only need to solve the following
\begin{align}
	(\beta^{(t)},\bm{g}^{(t)}) &= \argmin_{\beta,\bm{g}} \sum_{i=1}^N C(y_i)^{(t-1)} \exp \left(-\frac{1}{K} \mathbf{U}'_i (\bm{f}^{(t-1)}(\bm{x}_i) + \beta^{(t)} \bm{g}^{(t)}(\bm{x}_i))  \right) \nonumber\\
	&= \argmin_{\beta,\bm{g}} \sum_{i=1}^N \underbrace{C(y_i)^{(t-1)} \exp \left(-\frac{1}{K} \mathbf{U}'_i \bm{f}^{(t-1)}(\bm{x}_i) \right)}_{D_t(i)} \exp \left(-\frac{1}{K} \beta^{(t)} \mathbf{U}'_i \bm{g}^{(t)}(\bm{x}_i)  \right) \nonumber\\
	&= \argmin_{\beta,\bm{g}} \sum_{i=1}^N D_t(i) \exp \left(-\frac{1}{K} \beta^{(t)} \mathbf{U}'_i \bm{g}^{(t)}(\bm{x}_i)  \right), \label{eq:14}
\end{align}
where $D_t(i)$ does not depend on either $\beta$ or $\bm{g}(\bm{x})$ and is equivalent to the unnormalized distribution of observation weights in the $t$-th iteration in Algorithm \ref{alg:sammec2}. We notice that $\bm{g}^{(t)}(\bm{x})$ in (\ref{eq:14}) is a one-to-one correspondence with the multi-class classifier $h_t(\bm{x})$ in Algorithm \ref{alg:sammec2} in the following manner:
\[
h_t(\bm{x})=k, \qquad \text{if} \quad \bm{g}^{(t)}_k(\bm{x})=1.
\]
Therefore, in essence, solving for $\bm{g}^{(t)}(\bm{x})$ is equivalent to finding  $h_t(\bm{x})$ in Algorithm \ref{alg:sammec2}. 

\begin{proposition}
The solution to the optimization expressed as
\[
(\beta^{(t)},\bm{g}^{(t)}) = \argmin_{\beta,\bm{g}} \sum_{i=1}^N D_t(i) \exp \left(-\frac{1}{K} \beta^{(t)} \mathbf{U}'_i \bm{g}^{(t)}(\bm{x}_i)  \right)
\]
has the following form:
\begin{align}
	& \beta^{(t)}= \frac{(K-1)^2}{K} \left[ \log\left(\frac{1-\epsilon_t}{\epsilon_t}\right) + \log(K-1) \right] = \frac{(K-1)^2}{K} \alpha_t, \label{eq:15} \\
	& h_t(\bm{x}) = \argmin_h \sum_{i=1}^N D_t(i) I(y_i \ne h_t(\bm{x}_i)), \label{eq:16} 
\end{align}
where
\[
\epsilon_t = \frac{\sum_{i=1}^N D_t(i) I(y_i \ne h_t(\bm{x}_i))}{\sum_{i=1}^N D_t(i)}
\]
and
\[
\alpha_t = \log\left(\frac{1-\epsilon_t}{\epsilon_t}\right) + \log(K-1).
\]
\begin{proof}
To find $\bm{g}^{(t)}(\bm{x})$ in (\ref{eq:14}), first, we fix $\beta^{(t)}$. Let us consider the case where $\mathbf{U}_i \ne \bm{g}(\bm{x}_i)$. We have
\begin{equation}
	\mathbf{U}'_i \bm{g}(\bm{x}_i) = \frac{K-2}{(K-1)^2} - \frac{2}{K-1} = -\frac{K}{(K-1)^2}. \label{eq:17}
\end{equation}
On the other hand, when $\mathbf{U}_i = \bm{g}(\bm{x}_i)$, we have
\begin{equation}
	\mathbf{U}'_i \bm{g}(\bm{x}_i) = \sum_{j \ne k} \left(-\frac{1}{K-1}\right)^2 + 1 = \frac{K-1}{(K-1)^2} + 1 = \frac{K}{K-1}. \label{eq:18}
\end{equation}
Equations (\ref{eq:17}) and (\ref{eq:18}) lead us to
\begin{align}
	&\sum_{i=1}^N D_t(i) \exp \left(-\frac{1}{K} \beta^{(t)} \mathbf{U}'_i \bm{g}^{(t)}(\bm{x}_i)  \right)\nonumber\\
	= &\sum_{i=1}^N D_t(i) \exp \left(\frac{1}{K} \beta^{(t)} \-\frac{K}{(K-1)^2} I(y_i \ne h_t(\bm{x}_i)) \right) + \sum_{i=1}^N D_t(i) \exp \left(-\frac{1}{K}\beta^{(t)} \frac{K}{K-1} I(y_i = h_t(\bm{x}_i)) \right) \nonumber\\
	= & \sum_{i=1}^N D_t(i) \exp \left(\frac{\beta^{(t)}}{(K-1)^2} I(y_i \ne h_t(\bm{x}_i)) \right) + \sum_{i=1}^N D_t(i) \exp \left(- \frac{\beta^{(t)}}{K-1} I(y_i = h_t(\bm{x}_i)) \right)\nonumber\\
	= & e^{-\beta^{(t)}/(K-1)} \sum_{i=1}^N D_t(i)  + \left(e^{\beta^{(t)}/(K-1)^2} - e^{-\beta^{(t)}/(K-1)}\right) \sum_{i=1}^N D_t(i)  I(y_i \ne h_t(\bm{x}_i)) \label{eq:19}
\end{align}
From (\ref{eq:19}), since only the last sum depends on the classifier $h_t$, for a fixed value of $\beta$, solution to $\bm{g}^{(t)}$ results in:
\begin{equation*}
	h_t(\bm{x}) = \arg \min_h \sum_{i=1}^N D_t(i) I(y_i \ne h_t(\bm{x}_i)). 
\end{equation*}
Now let us find $\beta^{(t)}$ given $h_t$. If we  define $\epsilon_t$ as
\begin{equation}
	\epsilon_t = \frac{\sum_{i=1}^N D_t(i)I(y_i \ne h_t(\bm{x}_i))}{\sum_{i=1}^N D_t(i)}, \label{eq:20}
\end{equation}
plugging (\ref{eq:17}), (\ref{eq:18}), and (\ref{eq:20}) into (\ref{eq:14}), we would have
\[
	\sum_{i=1}^N D_t(i) \exp \left(-\frac{1}{K} \beta^{(t)}\mathbf{U}'_i \bm{g}^{(t)}(\bm{x}_i)  \right)= \sum_{i=1}^N D_t(i) \left[\left(e^{\beta^{(t)}/(K-1)^2} - e^{-\beta^{(t)}/(K-1)}\right) \epsilon_t + 
	e^{-\beta^{(t)}/(K-1)}\right].
\]
The summation component does not affect the minimization so that, differentiating with respect to $\beta^{(t)}$ and setting to zero, we get
\[
\left(\frac{1}{(K-1)^2} e^{\beta^{(t)}/(K-1)^2} + \frac{1}{K-1} e^{-\beta^{(t)}/(K-1)}\right) \epsilon_t - \frac{1}{K-1} e^{-\beta^{(t)}/(K-1)} = 0 
\]
and factoring out the term $(1/(K-1)) \exp(-\beta^{(t)}/(K-1))$, we get
\[
\frac{1}{K-1} e^{-\beta^{(t)}/(K-1)} \left[\left( \frac{1}{K-1} \exp\left(\frac{\beta^{(t)}}{(K-1)^2} + \frac{\beta^{(t)}}{K-1}\right) +1\right) \epsilon_t -1\right] = 0.
\]
Since we are minimizing a convex function of $\beta^{(t)}$, the optimal solution is
\[
\beta^{(t)}= \frac{(K-1)^2}{K} \left[ \log\left(\frac{1-\epsilon_t}{\epsilon_t}\right) + \log(K-1) \right] =\frac{(K-1)^2}{K} \alpha_t,
\]
\end{proof}
\end{proposition}

The terms $\epsilon_t$ and $\alpha_t$ are equivalent to those in Algorithm (\ref{alg:sammec2}). Subsequently, we can deduce the updating equation for the distribution of the observation weights in Algorithm (\ref{alg:sammec2}) after normalization.

\begin{proposition}
The distribution of the observation weights at each iteration simplifies to
\begin{equation}
\begin{cases}
D_{t}(i)C(y_i)\exp(-\alpha_t), & \text{for \ } \mathbf{U}_i =\bm{g}^{(t)}(\bm{x}_i) \\
D_{t}(i)C(y_i), & \text{for \ } \mathbf{U}_i \ne \bm{g}^{(t)}(\bm{x}_i), \\ 
\end{cases} \label{eq:21}
\end{equation}
Equation (\ref{eq:21}) is equivalent to the updating of the weights in Algorithm (\ref{alg:sammec2}) after normalization.
\begin{proof}
From equation (\ref{eq:13}), multiplying both sides by $-(1/K)\mathbf{U}_i$, exponentiating, and multiplying both sides by $C(y_i)^{(t)}$, we get
\[
	C(y_i)^{(t)} \exp\left(-\frac{1}{K} \mathbf{U}'_i \bm{f}^{(t)}(\bm{x}_i) \right) = C(y_i)^{(t-1)} \exp \left(-\frac{1}{K} \mathbf{U}'_i \bm{f}^{(t-1)}(\bm{x}_i) \right)C(y_i)\exp \left(-\frac{1}{K} \beta^{(t)} \mathbf{U}'_i \bm{g}^{(t)}(\bm{x}_i) \right)
\]
for which can be written as
\[
	D_{t+1}(i) = D_t(i) C(y_i)\exp \left(-\frac{1}{K}  \beta^{(t)} \mathbf{U}'_i \bm{g}^{(t)}(\bm{x}_i) \right),
\]
where we define
\[
D_{t+1}(i) = C(y_i)^{(t)} \exp\left(-\frac{1}{K} \mathbf{U}'_i \bm{f}^{(t)}(\bm{x}_i) \right)
\]
\[
D_t(i) = C(y_i)^{(t-1)} \exp\left(-\frac{1}{K}\mathbf{U}'_i\bm{f}^{(t-1)}(\bm{x}_i) \right).
\]
The weight at each iteration can further be simplified to
\begin{equation}
	D_t(i)C(y_i)\exp \left(-\frac{(K-1)^2}{K^2} \alpha_t \mathbf{U}'_i \bm{g}^{(t)}(\bm{x}_i) \right) 	=
	\begin{cases}
		D_t(i)C(y_i)\exp\left(- \frac{\alpha_t(K-1)}{K}\right), & \text{for } \mathbf{U}_i =\bm{g}^{(t)}(\bm{x}_i) \\
		D_t(i)C(y_i)\exp\left(\frac{\alpha_t}{K}\right), & \text{for } \mathbf{U}_i \ne \bm{g}^{(t)}(\bm{x}_i) \\ 
	\end{cases}\label{eq:22}
\end{equation}
Multiplying (\ref{eq:22}) by $\exp\left(\frac{-\alpha_t}{K}\right)$,
\[
\begin{cases}
D_t(i)C(y_i)\exp\left(- \frac{\alpha_t(K-1)}{K}\right)\exp\left( \frac{-\alpha_t}{K}\right)  =D_{t}(i)C(y_i)\exp(-\alpha_t), & \text{for \ } \mathbf{U}_i =\bm{g}^{(t)}(\bm{x}_i) \\
D_t(i)C(y_i)\exp\left(\frac{\alpha_t}{K}\right)\exp\left( \frac{-\alpha_t}{K}\right)  =  D_{t}(i)C(y_i), & \text{for \ } \mathbf{U}_i \ne \bm{g}^{(t)}(\bm{x}_i), \\ 
\end{cases}
\]
which proves our proposition. To show equivalence to the updating of the weights in Algorithm (\ref{alg:sammec2}), we note that the cases where $\mathbf{U}_i =\bm{g}^{(t)}(\bm{x}_i)$ and $\mathbf{U}_i \ne \bm{g}^{(t)}(\bm{x}_i)$ are equivalent to the cases where $y_i = h_t(\bm{x}_i)$ and $y_i \ne h_t(\bm{x}_i)$, respectively.
\end{proof}
\end{proposition}
It should be straightforward to show that the final classifier is the solution
\[
\argmax_{k}(f_1^{(T)}(\bm{x}_i),f_2^{(T)}(\bm{x}_i),\ldots,f_K^{(T)}(\bm{x}_i))',
\]
which is equivalent to
\begin{equation*}
	H(\bm{x}_i) = \argmax_{k} \sum_{t=1}^T \alpha_t I(\bm{g}^{(t)}(\bm{x}_i) = k).
\end{equation*}

\section{Numerical experiments} \label{sec:exp}

This section examines the differences between \texttt{SAMME.C2} and \texttt{SAMME} in how each model is trained, further exploring the superiority of \texttt{SAMME.C2} over \texttt{SAMME} in handling the issue regarding imbalanced data. To accomplish this, we make use of simulated dataset with a highly imbalanced three-class response variable. To generate such a simulated dataset, we utilize the \textit{Scikit-learn} Python module described in \citet{pedregosa2011scikit}. The \textpython{make\_classification} Application Programming Interface is employed with parameterization executed as

\begin{scriptsize}
	\begin{verbatim}
		"""Make Simulation"""
		from sklearn.datasets  import make_classification
		X, y = make_classification(n_samples=100000, n_features=50, n_informative=5-, n_redundant=0, n_repeated=0,
		n_classes=3, n_clusters_per_class=2, class_sep=2, flip_y=0, weights=[0.90,0.09,0.01], random_state=16)
	\end{verbatim}
\end{scriptsize}

\begin{figure}[htbp]
    \centering
    \includegraphics[width = \linewidth]{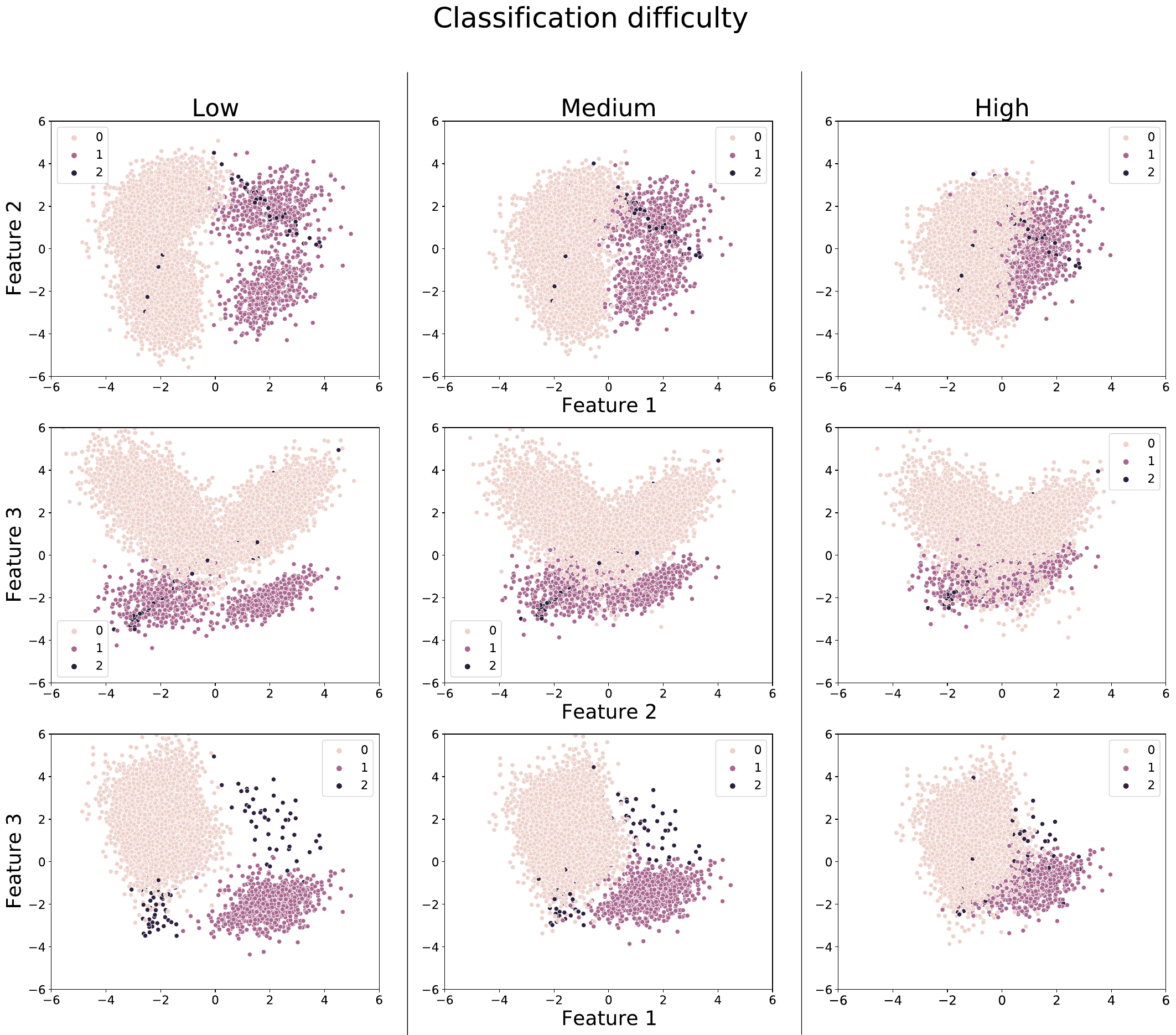} 
    \caption{Visualization of three synthetic datasets having 3 classes with different levels of classification difficulty.} \label{fig7:2D}
\end{figure}

This script generates 100,000 samples with 50 features and 3 classes, deliberately creating a highly imbalanced dataset by setting the ratio for each class as 90\%, 9\%, and 1\%, respectively. Changing the parameter of \textpython{class\_sep} adjusts the difficulty of the classification task. The samples no longer remain easily separable in the case of a lower value of \textpython{class\_sep}. To investigate and compare running processes of algorithms with different level of difficulty, three sets of data was created adjusting this parameter: for high classification difficulty, we set \textpython{class\_sep}=1, for medium classification difficulty, we set \textpython{class\_sep}=1.5, and for low classification difficulty, we set \textpython{class\_sep}=2. In Figure \ref{fig7:2D}, we visualize these three different difficulties of classification tasks, with each difficulty separated by columns. The figure clearly shows that low classification difficulty means the samples are easily separable; the opposite holds for high classification difficulty. For ease of visualization, we only use 3 features instead of the 50 features, and we exhibit a 3-dimensional data structure by pairing and drawing three 2-dimensional graphs. We kindly ask the reader to refer to the package for further explanation of the other input parameters. For training, we use 75\% of the data, and the rest are used for testing.

For classification problems, the most common performance statistics is accuracy, which is the proportion of all observations that were correctly classified. For obvious reasons, this is an irrelevant measure for imbalanced datasets. As alternative statistics, we consider Recall to measure the performance.  The Recall statistics, sometimes called the sensitivity, for class $i$, $R_i$, is defined to be the proportion of observations in class $i$ correctly classified. It has been discussed (\cite{fernandez2018learning}) that the Recall, or sensitivity, is usually a more interesting measure for imbalanced classification. To provide a single measure of performance for a given classifier, we use the geometric average of Recall statistics, denoted as \text{MAvG}, as follows:
\begin{equation}
\text{MAvG} = (R_1 \times R_2 \times \ldots \times R_K)^{1/K}. \label{eq:mavg}
\end{equation}
It is straightforward to show that when we take the log of both sides of this performance metric, we get an average of the log of all the Recall statistics. This log transformation leads us to a metric that provides for impartiality of the importance of accurately classifying observations for all classes. In the case of severely imbalanced datasets, the MAvG metric allows us to correctly classify more observations of the minority class while sacrificing misclassifications of observations in the majority class. In effect, the MAvG metric is a sensible performance measure for severely imbalanced datasets; this performance metric  is also used as the criterion for the hyperparameter optimization in GA to determine the cost values used in the \texttt{SAMME.C2} algorithm. The concept of MAvG used for imbalanced datasets originated from the work of \citet{fowlkes1983gmean}.

\begin{figure}[h!]
	\centering
	\includegraphics[width=3.2 in ]{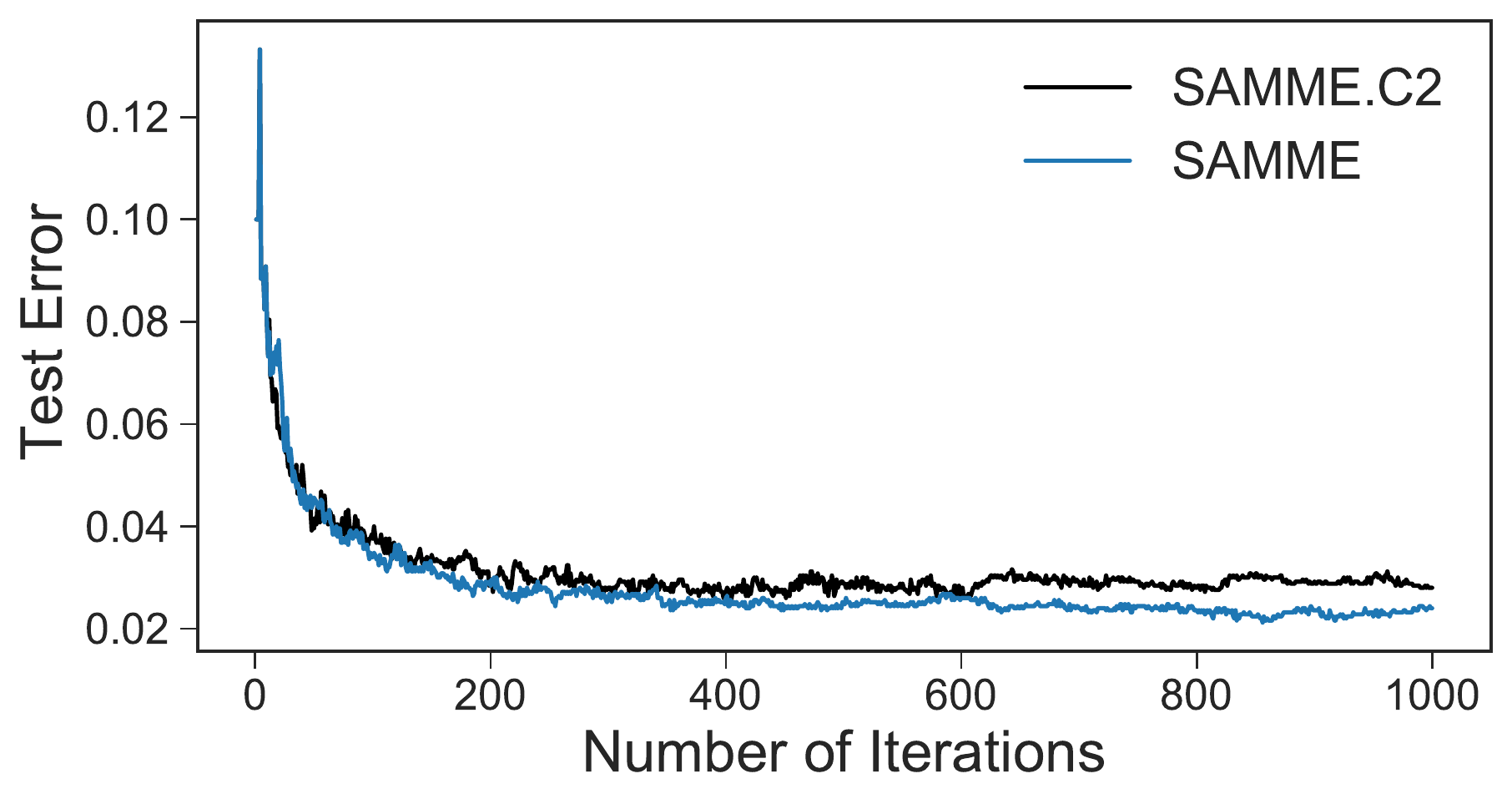} 
	\includegraphics[width=3.2 in]{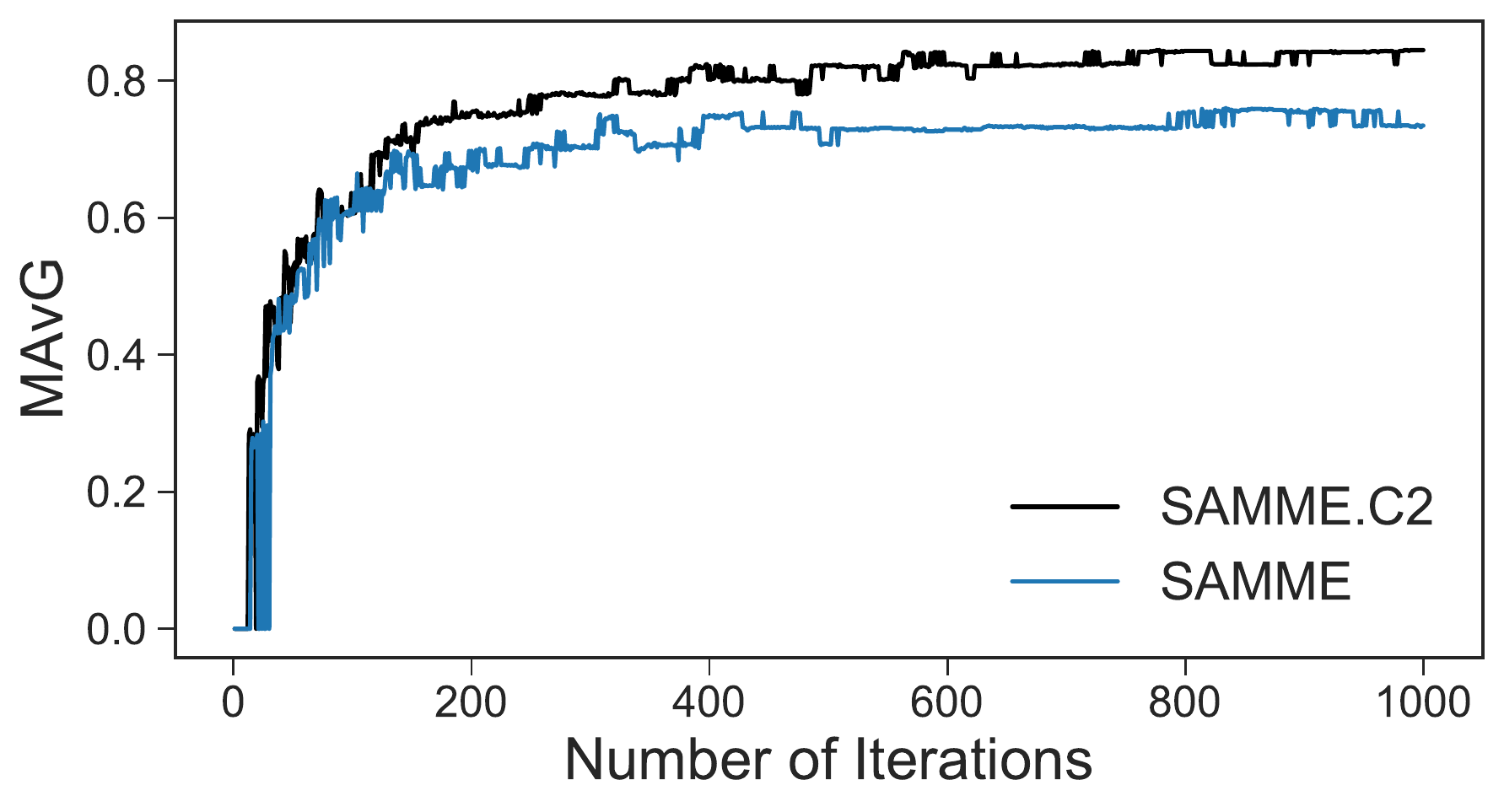} 
	\caption{Comparison of test error and test MAvG between \texttt{SAMME.C2} and \texttt{SAMME} with 1,000 decision stumps using the dataset of low level of classification difficulty.} \label{fig1:testerror2}
\end{figure}

\begin{figure}[h!]
	\centering
	\includegraphics[width=3.2 in ]{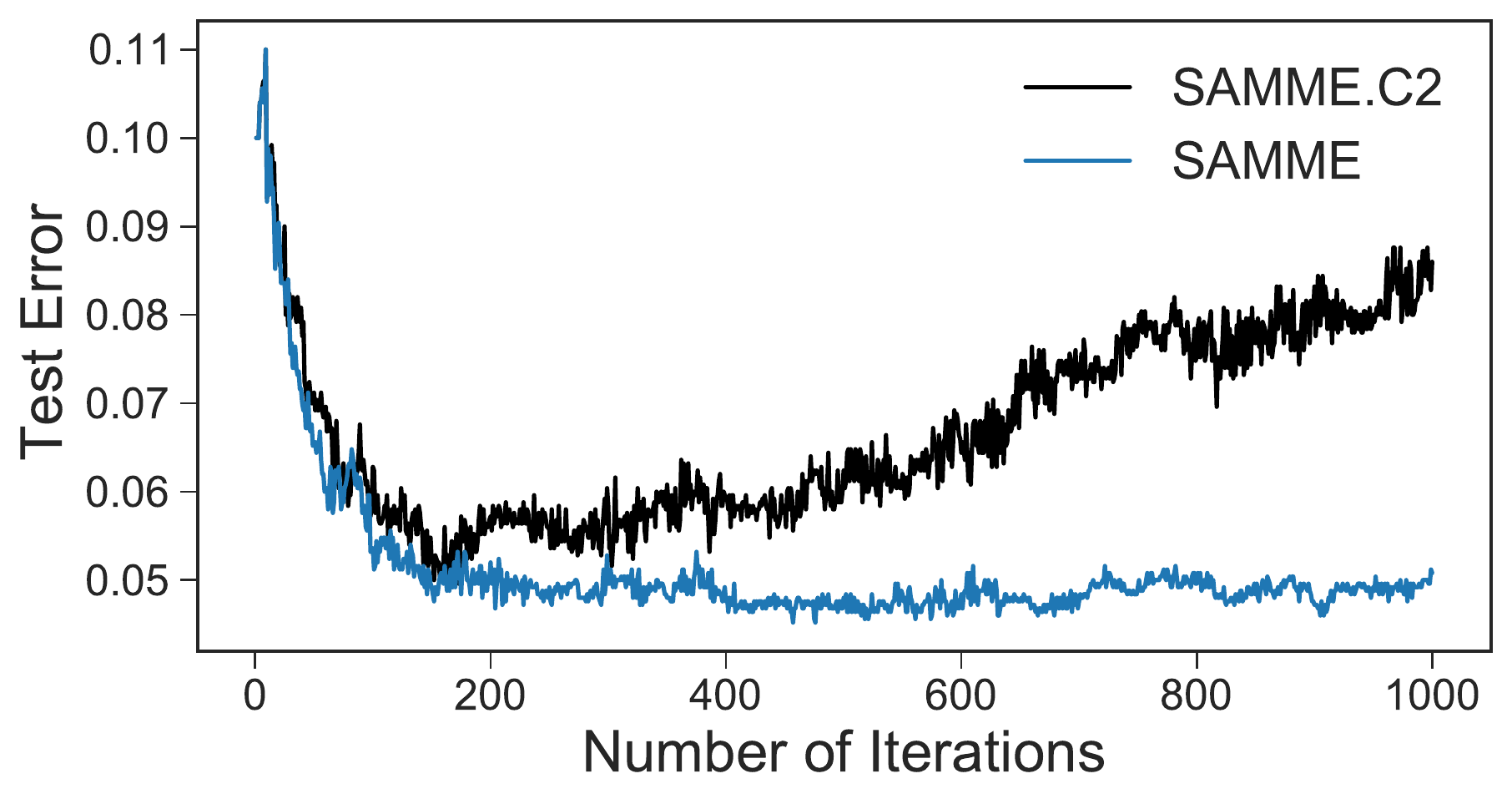} 
	\includegraphics[width=3.2in]{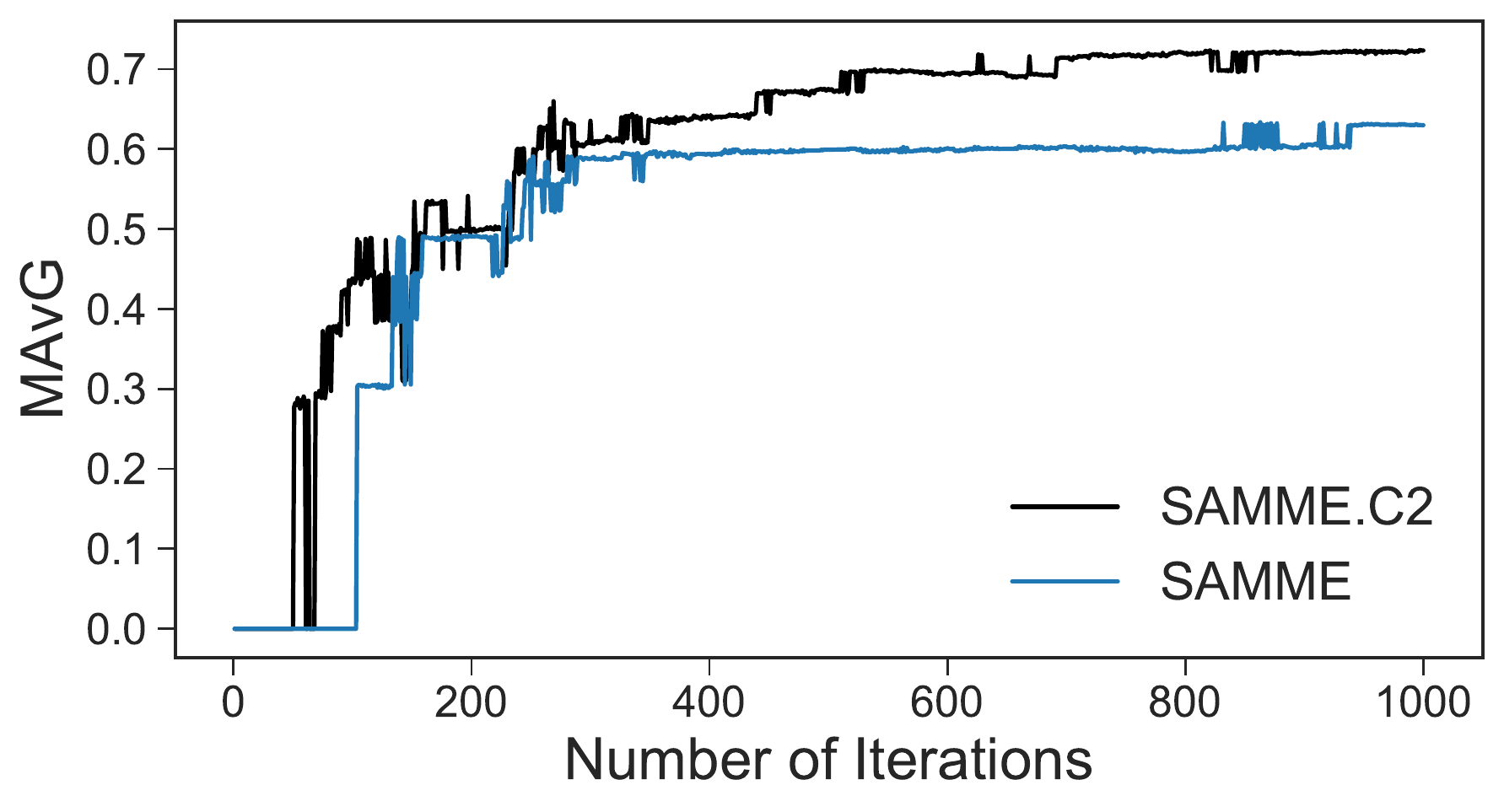} 
	\caption{Comparison of test error and test MAvG between \texttt{SAMME.C2} and \texttt{SAMME} with 1,000 decision stumps using the dataset of medium level of classification difficulty.} \label{fig2:testerror15}
\end{figure}

\begin{figure}[h!]
 	\centering
 	\includegraphics[width=3.2 in ]{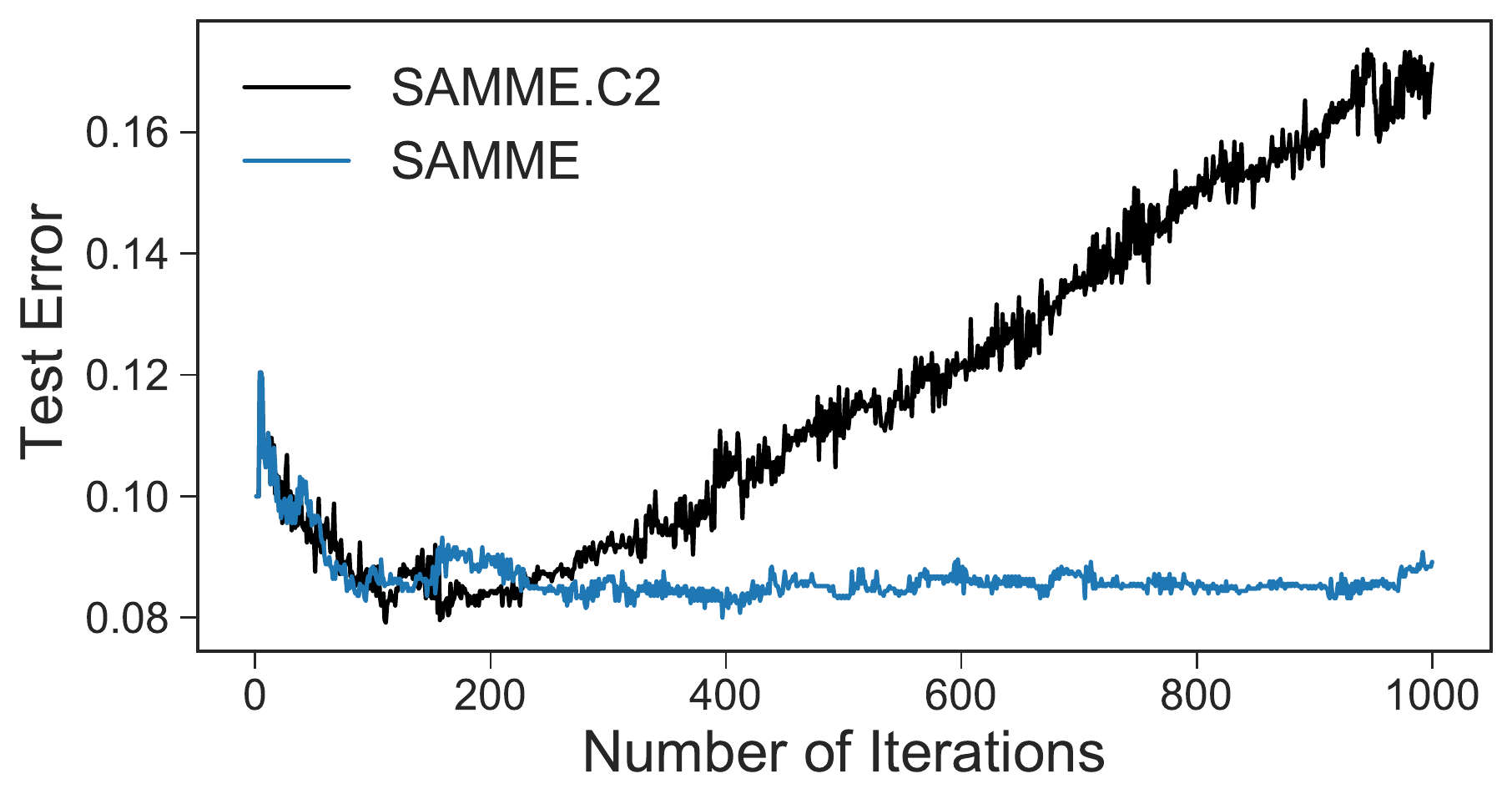} 
 	\includegraphics[width=3.2 in]{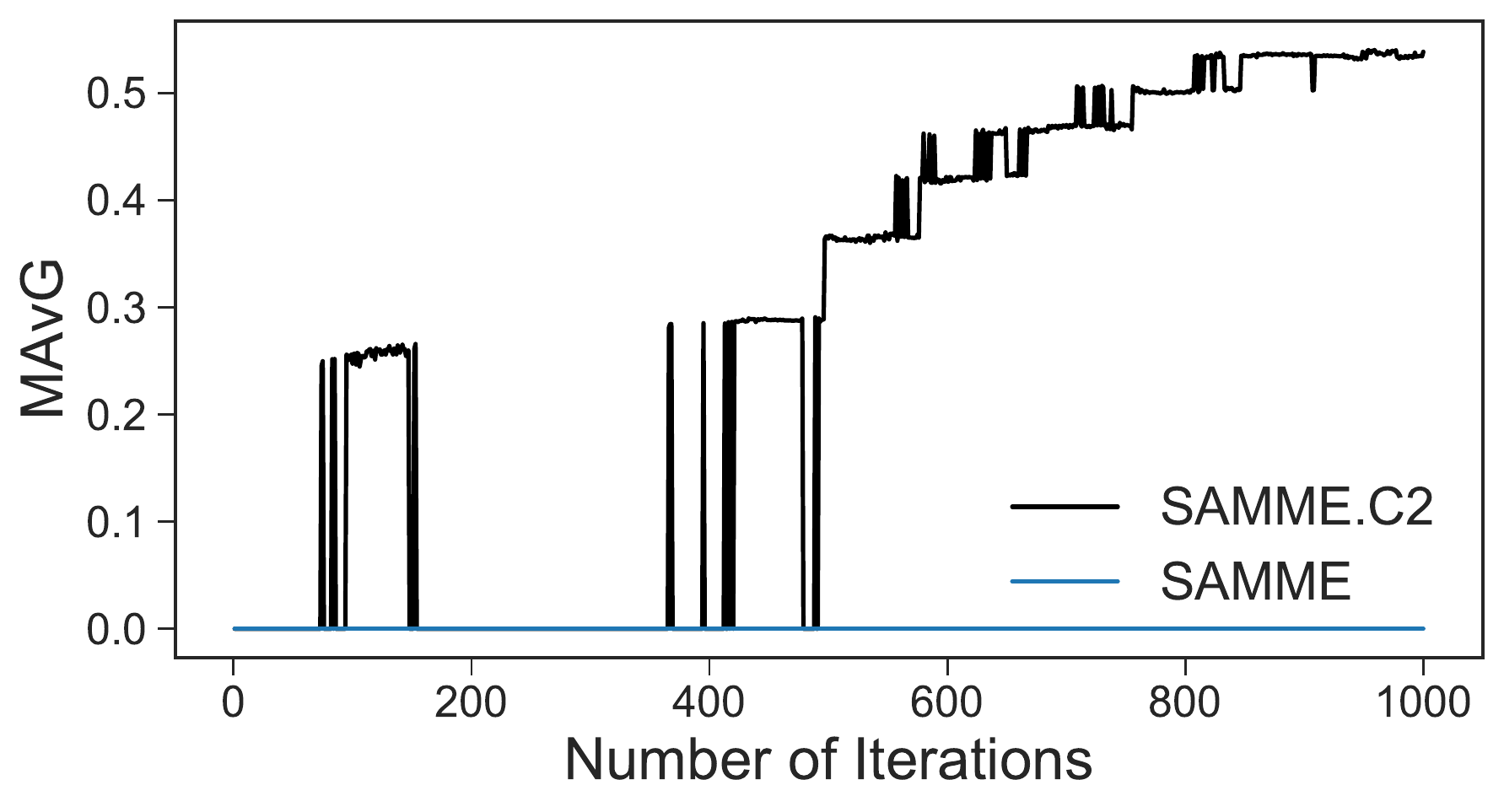} 
 	\caption{Comparison of test error and test MAvG between \texttt{SAMME.C2} and \texttt{SAMME} with 1,000 decision stumps using the dataset of high level of classification difficulty.} \label{fig3:testerror1}
\end{figure}

To examine running processes of \texttt{SAMME.C2} and \texttt{SAMME}, each algorithm with 1,000 decision stumps is trained using three datasets. The decision stump is a decision tree with one depth, which plays the role of a weak learner in the algorithms.  Figures \ref{fig1:testerror2}, \ref{fig2:testerror15}, and \ref{fig3:testerror1} show the resulting test errors and test MAvG of  \texttt{SAMME.C2} and \texttt{SAMME}, after training newly added decision trees using the datasets of varying (low, medium, and high) level of difficulty. All figures are produced for increasing number of iterations, with each iteration referring to new decision stump.


With \texttt{SAMME} algorithm, the objective is to reduce the test error, the misclassification rate. Therefore, when model is trained with severely imbalanced data, it puts more weight on a majority class since the majority class can significantly reduce the test error. For example, based on the simulated datasets in these numerical experiments, a model can be constructed assigning all observations in the majority class. In which case, we will get a misclassification rate of 10\% which can be deemed small. Therefore, the test error is not a meaningful performance metric for severely imbalanced datasets. All figures show small test errors for \texttt{SAMME} algorithm, but when the \texttt{SAMME.C2} algorithm is used, test errors are clearly low for low level of difficulty of classification and rapidly becomes worst for very high level of difficulty of classification.

On the other hand, all three figures show that \texttt{SAMME.C2} algorithm produces better MAvG performance metric for various level of difficulty of classification. It is noted further than in the case when we have a high level of difficulty of classification, the \texttt{SAMME.C2} produces a much improved MAvG metric than the \texttt{SAMME} algorithm. This results in spite of the worst test errors. This leads us to infer that in order to have a higher accuracy for minority class, \texttt{SAMME.C2} has to sacrifice accuracy for majority class. This becomes clearer in the subsequent figure.

\begin{figure}[htbp]
	\centering
	\includegraphics[width=6.4 in]{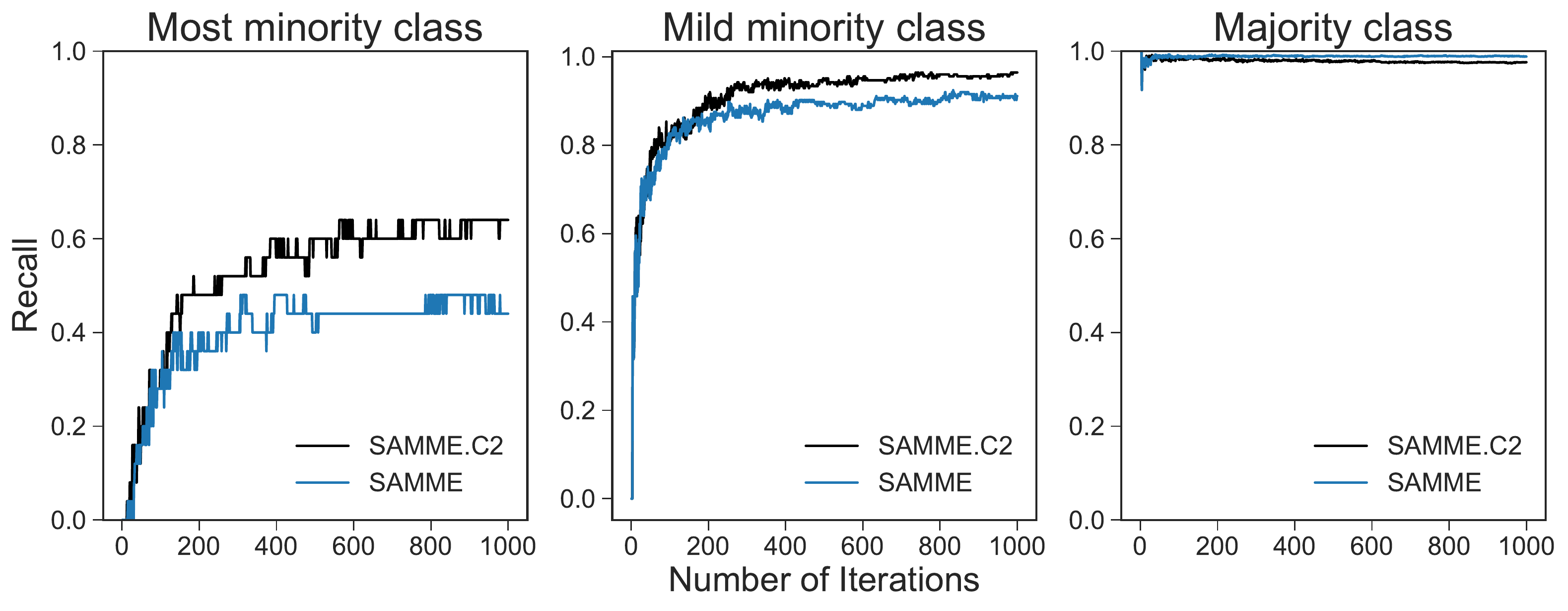} 
	\includegraphics[width=6.4 in]{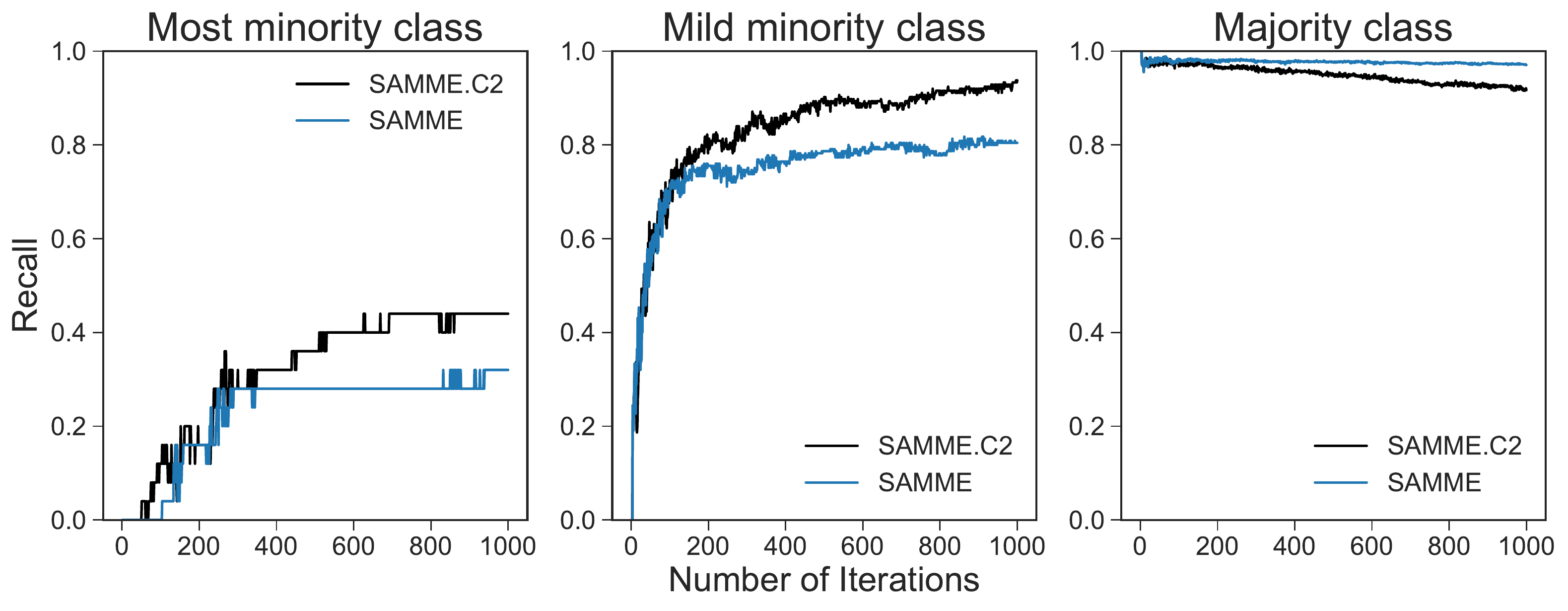} 
	\includegraphics[width=6.4 in]{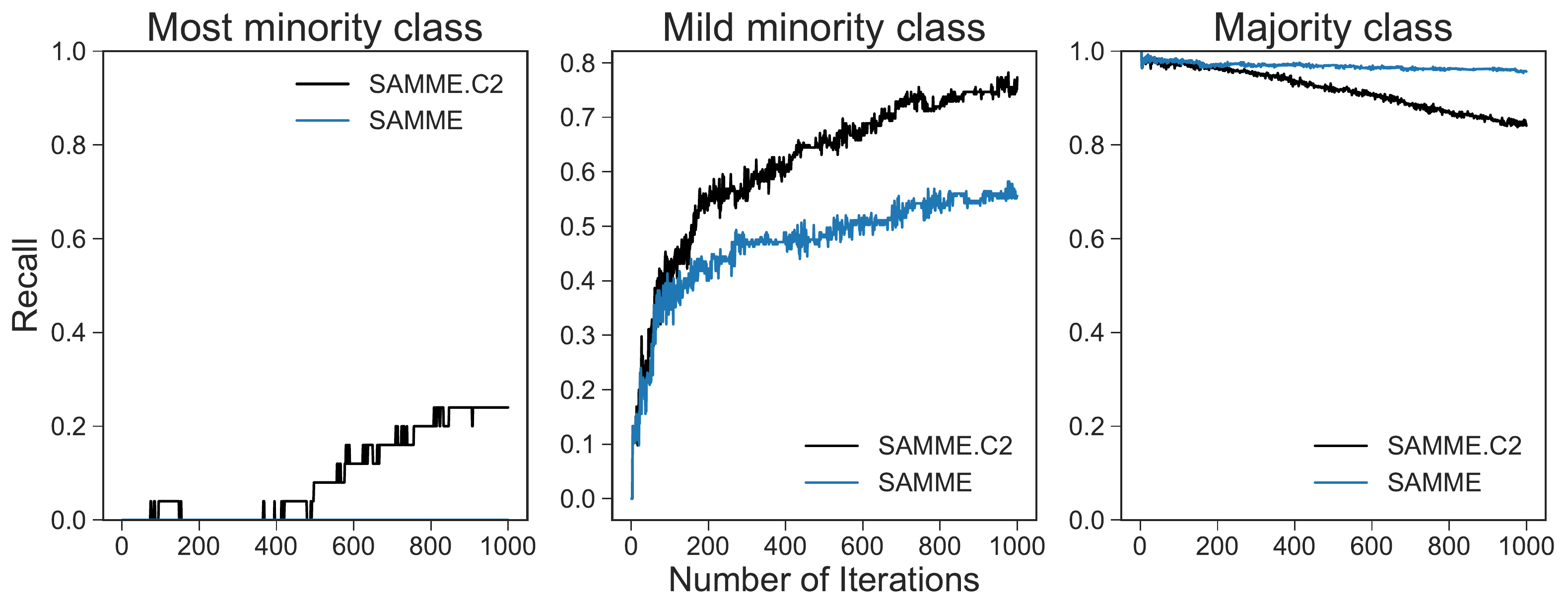} 
	\caption{Comparison of Recall statistics of each class between \texttt{SAMME.C2} and \texttt{SAMME} with 1,000 decision stumps using the dataset of low level (Top), medium level (Middle) and high level (Bottom) difficulty of classification.} \label{fig5:ga}
\end{figure}

In Figure \ref{fig5:ga}, we can observe the mechanism of \texttt{SAMME.C2} in more details by examining Recall statistics of each of the three classes. Regardless of the complexity of the classification task, \texttt{SAMME.C2} classifies minority classes much more accurately than \texttt{SAMME}. However, accuracy from minority classes is gained by sacrificing the accuracy of majority class. In other words, the primary difference between \texttt{SAMME.C2} and \texttt{SAMME} occurs based on whether the model is trained focusing on reducing test errors or improving a more balanced accuracy of classification across all classes. Apparently, as the level of classification task increases, Figure \ref{fig5:ga} shows that, to correctly classify observations in the minority class, \texttt{SAMME.C2} has to correspondingly reduce accuracy of observations in the majority class. This is a very important result because when observations in the minority class for severely imbalanced datasets are extremely difficult to classify, \texttt{SAMME} assigns nearly all observations in the majority class. Differently said, \texttt{SAMME} assigns nearly no observation in the minority class.

\begin{figure}[htbp]
\centering
\includegraphics[width=5 in ]{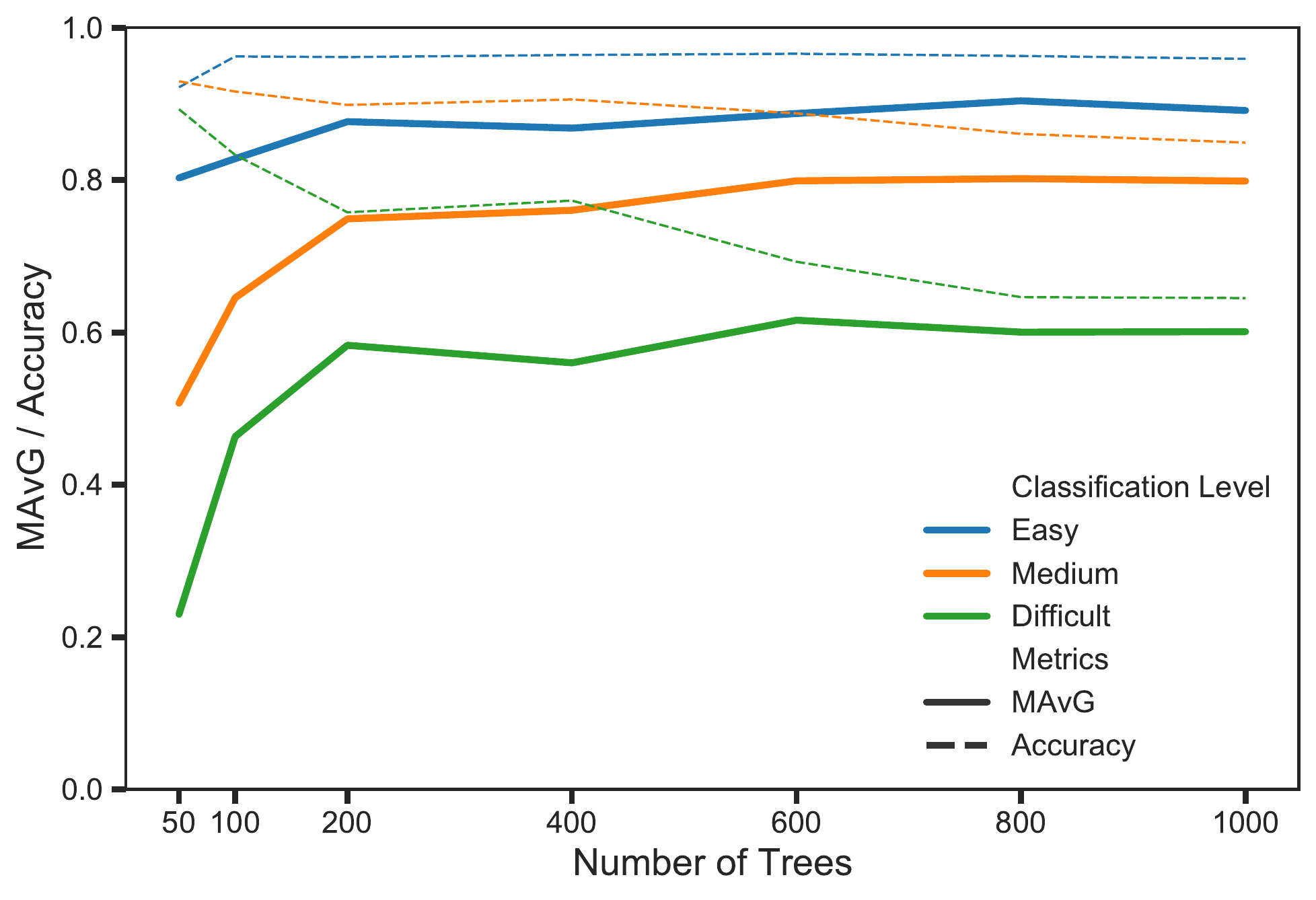} 
\caption{5-fold cross validation of MAvG and accuracy of \texttt{SAMME.C2} with adjustments of the number of decision stumps (trees).} \label{fig4:metric}
\end{figure}

The number of times we iterate to reach an optimal classifier is clearly directly linked to the number of decision stumps we use as weak learners. The more weak learners we use the closer we can reach a desired convergence of our MAvG performance metric. In essence, this can impact the computational efficiency of our iterative algorithm. To do this investigation, we examine for a reasonable number of decision stumps to use for the \texttt{SAMME.C2} algorithm by exploring the change in the value of MAvG vis-a-vis the number of decision stumps. Figure \ref{fig4:metric} exhibits the results of this investigation.

In the figure, for each level of difficulty of the classification tasks, we examine how changing the number of trees affects reaching the optimal MAvG performance metric with training. The figure shows the effects for various levels of difficulty of classification, varying the number of decision stumps or trees from 50, 100, and intervals of 100 up to 1000. For each time we train a model, the cost values are newly tuned through Genetic Algorithm. The results in Figure \ref{fig4:metric} exhibit solid lines determined according to 5-fold cross validation MAvGs. For reference purposes, we also give the corresponding 5-fold cross validation accuracy values shown as dashed lines. For all levels of difficulty, MAvGs increase sharply before 200 decision stumps, however, after that, MAvGs do not improve significantly with increasing number of decision stumps. Therefore, we conclude that at least 200 decision stumps are necessary for \texttt{SAMME.C2} to perform suitably and favorably.

\begin{figure}[h!]
	\centering
	\includegraphics[width=6.5 in]{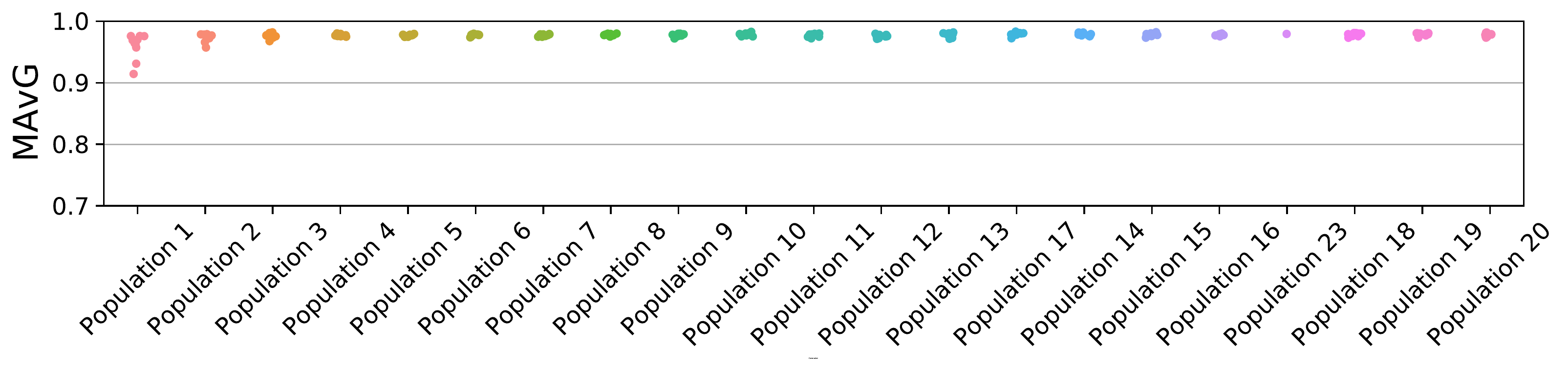} 
	\includegraphics[width=6.5 in]{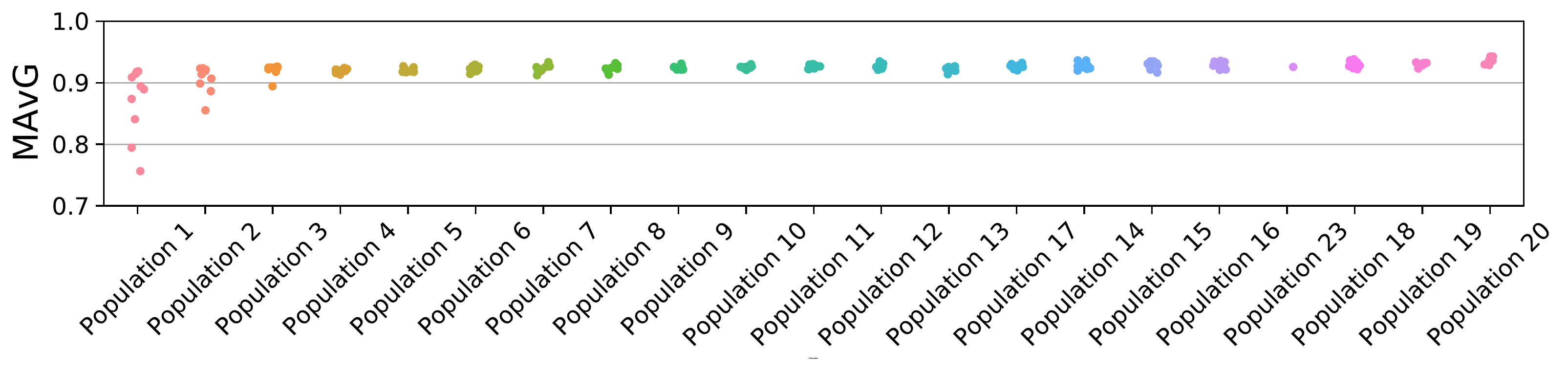} 
	\includegraphics[width=6.5 in]{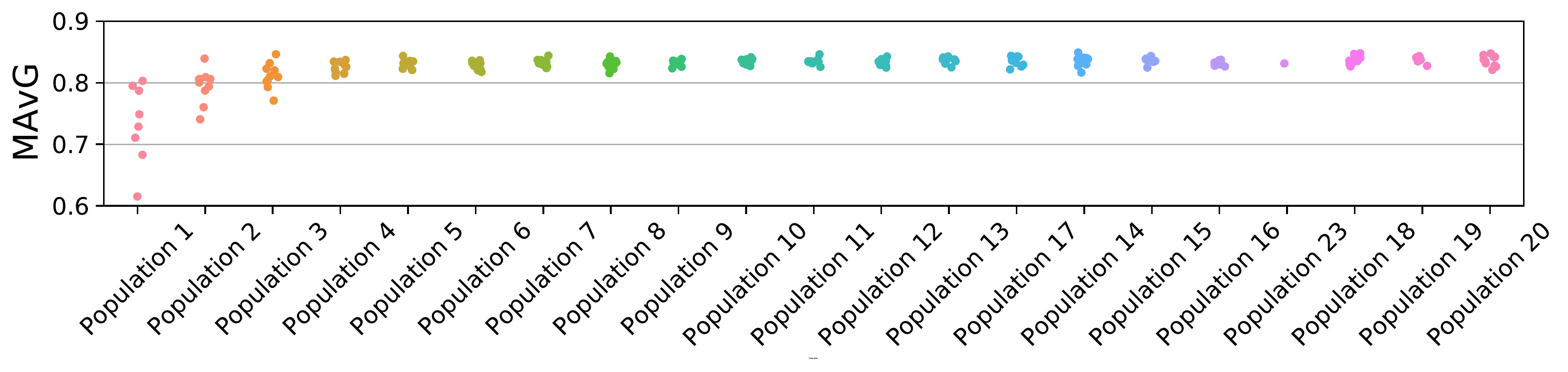} 
	\caption{Number of populations in the Genetic Algorithm (GA) to tune the cost values for each class, according to level of difficulty: top (low level), middle (medium level), and bottom (high level).} \label{fig6:ga}
\end{figure}

Finally, we examine the proper number of populations ($P$) in GA explained in Section \ref{sub.cost} to tune the cost values of each class for \texttt{SAMME.C2}.  To narrow the possible interval of selecting values, the cost value of the most minority class is fixed at 0.999. Since we should give the largest cost to the most minority class, obviously, cost values for other classes should be between 0 and 0.999. It has been demonstrated by initial experiments that, when we run \texttt{SAMME.C2} with over 200 decision stumps, the best cost values chosen from GA are in between 0.95 and 0.999. Based on these results, we determine the optimal cost values by choosing from the interval (0.95, 0.999) and we allow for randomness of around 0.001, in the mutation step of GA. Figure \ref{fig6:ga} reveals 10 values of MAvG according to 10 cost values in each population. As explained in section \ref{sub.cost}, the set of 10 cost values of each population is determined by the 10 MAvG values calculated with trained \texttt{SAMME.C2} using the set of 10 cost values of the previous population. For all three levels of classification difficulties, we arrive at the best cost values rather quite rapidly. We observe that just after the 4th population, the largest MAvG for each population is nearly similar. The assignment of cost values in \texttt{SAMME.C2} does not slow the overall estimation and training of the \texttt{SAMME.C2} algorithm.

We have used numerical experiments to have a better understanding of the \texttt{SAMME.C2} especially when compared to the \texttt{SAMME} algorithm. We find that \texttt{SAMME.C2} provides us a much more superior algorithm for learning and understanding observations in the minority class, regardless of the level of difficulty of classification embedded in the data. We also examined how \texttt{SAMME.C2} performs relative to other algorithms that handle severely imbalanced classes based on insurance telematics data. See \citet{so2021cost}.

\section{Concluding remarks} \label{sec:conclude}

Because of its potential use in a vast array of disciplines, classification predictive modeling will continue to be an important toolkit in machine learning. One of the most challenging aspects of classification task is finding an optimal procedure to handle observational data with skewed distribution across several classes. We find that there is now a growing body of literature that deals with real world classification tasks related to highly imbalanced multi-class problems. In spite of this growing demand, there is insufficient work on methods to handle severely imbalanced data in a multi-class classification.

In this paper, we presented what we believe is a promising algorithm for handling severely imbalanced multi-class classification. The proposed method, which we refer to as \texttt{SAMME.C2}, combines the benefits of iterative learning from weak learners through the AdaBoost scheme and increased repeated learning of observations in the minority class through a cost-sensitive learning scheme. We provided a mathematical proof that the optimal procedure resulting in \texttt{SAMME.C2} is equivalent to an additive model with a minimization of a multi-class cost-sensitive exponential loss function. The algorithm therefore belongs to the traditional statistical family of forward stagewise additive models. We additionally showed that based on the same multi-class cost-sensitive exponential loss function, \texttt{SAMME.C2} is an optimal Bayes classifier.

In order to expand our insights into \texttt{SAMME.C2} relative to \texttt{SAMME}, our numerical experiments are based on understanding the resulting differences when differing levels of difficulty in classification task is used. We therefore synthetically generated three simulated datasets that are distinguished according to these degrees of difficulty of classification. First, we note that the use of straightforward misclassification, or test errors, does not work well for severely imbalanced datasets. As has been proposed in the literature, the use of MAvG, a geometric average of recall statistics for all classes, is a more rational performance metric as it gives emphasis on being able to train and learn well from observations that belong to the more minority classes. By recording and tracking test errors, MAvGs, and recall statistics, the results of our numerical experiments reveal the superiority of \texttt{SAMME.C2} in classifying objects that belong to the minority class, regardless of the degree of difficulty of classification. This is at the little expense of sacrificing recall statistics for the majority class. For \texttt{SAMME.C2}, the recall statistics of minority classes are much more improved at each iteration than those of \texttt{SAMME}, but \texttt{SAMME.C2} has lower recall statistics for majority classes at all iterations than those of \texttt{SAMME}. We also showed the computational efficiency of \texttt{SAMME.C2} by investigating the most optimal number of weak learners, or iterations, in order to reach convergence. Based on our analysis, training as little as 200 decision stumps as weak learners can rationally stop the iteration.

\newpage

\section*{Appendix A. Detailed steps of the various algorithms}\label{appendix_a}
\addcontentsline{toc}{chapter}{Appendix}

\addcontentsline{toc}{section}{A. Detailed steps of the \texttt{AdaBoost.M1}, \texttt{SAMME} and \texttt{Ada.C2}}

\begin{algorithm}[h!]
\LinesNumberedHidden
\SetAlgoNoLine
	\KwData{$\ \bm{x}_i \in X$, $y_i \in Y = \{0,1\}$}
	\KwIn{$T$}
	\KwOut{Final classifier $\ H(\bm{x}_i)=\underset{k}{\mathrm{argmax}} {\  \sum_{t=1}^{T} \alpha_t I(h_t(\bm{x}_i) = k)}$}
	Set initial sample weights equally distributed:  $D_1(i) = \frac{1}{N}, \quad i =1,2,\ldots,N$ \;
	\For{$t=1, \ldots, T$}{
		Train weak classifier using the distribution $D_t$ \;
		Get weak classifier $h_t: X \rightarrow k \in \{0,1\}$ \;
		Compute error rate $\epsilon_t = \dfrac{\sum_{i=1}^{N} D_t(i) I(y_i \ne h_t(\bm{x}_i))}{\sum_{i=1}^N D_t(i)}$ \;
		Calculate weight $\alpha_t = \log\!\Big(\dfrac{1-\epsilon_t}{\epsilon_t}\Big)$ \;
		Update sample weights $D_{t+1}(i) = \dfrac{D_t(i) \exp(-\alpha_t I(y_i = h_t(\bm{x}_i)))}{\sum_{j=1}^{N} D_t(j) \exp(-\alpha_t I(y_j = h_t(\bm{x}_j)))}$ \;
	}
	\caption{\texttt{AdaBoost.M1}}\label{alg:ada.m1}
\end{algorithm}

\bigskip

\begin{algorithm}[h!]
\LinesNumberedHidden
\SetAlgoNoLine
	\KwData{$\ \bm{x}_i \in X$, $y_i \in Y = \{0,1\}$}
	\KwIn{$C(y_i) \in (0,1]$, $T$}
	\KwOut{Final classifier $\ H(\bm{x}_i) = \underset{k}{\mathrm{argmax}} {\  \sum_{t=1}^{T} \alpha_t I(h_t(\bm{x}_i) = k)}$}
	Set initial sample weights equally distributed:  $D_1(i) = \frac{1}{N}, \quad i =1,2,\ldots,N$ \;
	\For{$t=1, \ldots, T$}{
		Train weak classifier using the distribution $D_t$ \;
		Get weak classifier $h_t: X \rightarrow k \in \{0,1\}$ \;
		Compute error rate $\epsilon_t = \dfrac{\sum_{i=1}^{N} C(y_i)D_t(i) I(y_i \ne h_t(\bm{x}_i))}{\sum_{i=1}^N C(y_i)D_t(i)}$\;
		Choose weight $\alpha_t = \frac{1}{2} \log\!\Big(\dfrac{1-\epsilon_t}{\epsilon_t}\Big)$ \;
		Update sample weights $D_{t+1}(i) = \dfrac{C(y_i) \, D_t(i) \exp(-\alpha_t I(y_i = h_t(\bm{x}_i)))}{\sum_{j=1}^{N} C(y_j) \, D_t(j) \exp(-\alpha_t I(y_j = h_t(\bm{x}_j)))}$ \;
	}
	\caption{\texttt{Ada.C2}: cost-sensitive binary AdaBoost}\label{alg:adac2}
\end{algorithm}

\bigskip

\begin{algorithm}[h!]
\LinesNumberedHidden
\SetAlgoNoLine
	\KwData{$\ \bm{x}_i \in X$, $y_i \in Y = \{1,2,\ldots,K\}$}
	\KwIn{$T$}
	\KwOut{Final classifier $\ H(\bm{x}_i) = \underset{k}{\mathrm{argmax}} {\  \sum_{t=1}^{T} \alpha_t I(h_t(\bm{x}_i) = k)}$}
	Set initial sample weights equally distributed:  $D_1(i) = \frac{1}{N}, \quad i =1,2,\ldots,N$ \;
	\For{$t=1, \ldots, T$}{
		Train weak classifier using the distribution $D_t$ \;
		Get weak classifier $h_t: X \rightarrow k \in \{1,2,\ldots,K\}$ \;
		Compute error rate $\epsilon_t = \dfrac{\sum_{i=1}^{N} D_t(i) I(y_i \ne h_t(\bm{x}_i))}{\sum_{i=1}^N D_t(i)}$ \;
		Choose weight $\alpha_t = \log\!\Big(\dfrac{1-\epsilon_t}{\epsilon_t}\Big) + \log(K-1)$ \;
		Update sample weights $D_{t+1}(i) = \dfrac{D_t(i) \exp(-\alpha_t I(y_i = h_t(\bm{x}_i)))}{\sum_{j=1}^{N} D_t(j) \exp(-\alpha_t I(y_j = h_t(\bm{x}_j)))}$ \;
	}
	\caption{\texttt{SAMME}: multi-class AdaBoost}\label{alg:samme}
\end{algorithm}

\bigskip

\begin{algorithm}[H]
\LinesNumberedHidden
\SetAlgoNoLine
	\KwData{$\ \bm{x}_i \in X$, $y_i \in Y = \{1,2,\ldots,K\}$}
	\KwIn{$C(y_i) \in (0,1]$, $T$}
	\KwOut{Final classifier $\ H(\bm{x}_i)= \underset{k}{\mathrm{argmax}} {\  \sum_{t=1}^{T} \alpha_t I(h_t(\bm{x}_i) = k)}$}
	Set initial sample weights:  $D_1(i) = \frac{1}{N}, \quad i =1,2,\ldots,N$ \;
	\For{$t=1, \ldots, T$}{
		Train weak classifier using the distribution $D_t$\;
		Get weak classifier $h_t: X \rightarrow k \in \{1,2,\ldots,K\}$\; 
		Compute error rate  $\epsilon_t = \dfrac{\sum_{i=1}^{N} D_t(i) I(y_i \ne h_t(\bm{x}_i))}{\sum_{i=1}^N D_t(i)}$ \;
		Calculate weight $\alpha_t = \log\!\Big(\dfrac{1-\epsilon_t}{\epsilon_t}\Big) + \log(K-1)$ \;
		Update sample weights $D_{t+1}(i) = \dfrac{C(y_i) \, D_t(i) \exp(-\alpha_t I(y_i = h_t(\bm{x}_i)))}{\sum_{j=1}^{N} C(y_j) \, D_t(j) \exp(-\alpha_t I(y_j = h_t(\bm{x}_j)))}$ \;
	}
	\caption{\texttt{SAMME.C2}: cost-sensitive multi-class AdaBoost}\label{alg:sammec2}
\end{algorithm}

\bigskip

\bibliographystyle{apalike}
\bibliography{sammec2_theory_draft.bib}

\begin{thebibliography}{}

\bibitem[Bergstra et~al., 2011]{bergstra2011smbo}
Bergstra, J., Bardenet, R., Bengio, Y., and K{\'e}gl, B. (2011).
\newblock Algorithms for hyper-parameter optimization.
\newblock {\em Advances in neural information processing systems}, 24.

\bibitem[Bergstra and Bengio, 2012]{bergstra2012randomsearch}
Bergstra, J. and Bengio, Y. (2012).
\newblock Random search for hyper-parameter optimization.
\newblock {\em Journal of Machine Learning Research}, 13:281–305.

\bibitem[Chawla et~al., 2002]{chawla2002smote}
Chawla, N.~V., Bowyer, K.~W., Hall, L.~O., and Kegelmeyer, W.~P. (2002).
\newblock {SMOTE}: Synthetic minority over-sampling technique.
\newblock {\em Journal of Artificial Intelligence Research}, 16:321--357.

\bibitem[Dewi et~al., 2017]{dewi2017multiclass}
Dewi, F.~K., Fadhlurrahman, M. M.~R., Rahmanianto, M.~D., and Mahendra, R.
  (2017).
\newblock Multiclass sms message categorization: Beyond spam binary
  classification.
\newblock In {\em 2017 International Conference on Advanced Computer Science
  and Information Systems (ICACSIS)}, pages 210--215. IEEE.

\bibitem[Fern{\'a}ndez et~al., 2018]{fernandez2018learning}
Fern{\'a}ndez, A., Garc{\'\i}a, S., Galar, M., Prati, R.~C., Krawczyk, B., and
  Herrera, F. (2018).
\newblock {\em Learning from imbalanced data sets}, volume~11.
\newblock Springer.

\bibitem[Ferreira and Figueiredo, 2012]{ferreira2012review}
Ferreira, A.~J. and Figueiredo, M.~A. (2012).
\newblock Boosting algorithms: {A} review of methods, theory, and applications.
\newblock In Zhang, C. and Ma, Y., editors, {\em Ensemble Machine Learning:
  Methods and Applications}, chapter~2, pages 35--85. Springer Science.

\bibitem[Fowlkes and Mallows, 1983]{fowlkes1983gmean}
Fowlkes, E.~B. and Mallows, C. (1983).
\newblock A method for comparing two hierarchical clusterings.
\newblock {\em Journal of the American Statistical Association},
  78(383):553--569.

\bibitem[Freund and Schapire, 1997]{freund1997decision}
Freund, Y. and Schapire, R.~E. (1997).
\newblock A decision-theoretic generalization of on-line learning and an
  application to boosting.
\newblock {\em Journal of Computer and System Sciences}, 55(1):119--139.

\bibitem[Friedman et~al., 2000]{friedman2000addlog}
Friedman, J., Hastie, T., and Tibshirani, R. (2000).
\newblock Additive logistic regression: {A} statistical view of boosting.
\newblock {\em The Annals of Statistics}, 28(2):337--407.

\bibitem[Han et~al., 2019]{han2019fault}
Han, S., Choi, H.-J., Choi, S.-K., and Oh, J.-S. (2019).
\newblock Fault diagnosis of planetary gear carrier packs: A class imbalance
  and multiclass classification problem.
\newblock {\em International Journal of Precision Engineering and
  Manufacturing}, 20(2):167--179.

\bibitem[Hastie et~al., 2009]{hastie2009}
Hastie, T., Tibshirani, R., and Friedman, J. (2009).
\newblock {\em The Elements of Statistical Learning: Data Mining, Inference,
  and Prediction}.
\newblock Springer: New York.

\bibitem[Holland, 1975]{holland1975}
Holland, J.~H. (1975).
\newblock {\em Adaptation in Natural and Artifical Systems}.
\newblock Univesity of Michigan Press: Ann Arbor.

\bibitem[Jeong et~al., 2020]{jeong2020comparison}
Jeong, B., Cho, H., Kim, J., Kwon, S.~K., Hong, S., Lee, C., Kim, T., Park,
  M.~S., Hong, S., and Heo, T.-Y. (2020).
\newblock Comparison between statistical models and machine learning methods on
  classification for highly imbalanced multiclass kidney data.
\newblock {\em Diagnostics}, 10(6):415.

\bibitem[Kim et~al., 2016]{kim2016detecting}
Kim, Y.~J., Baik, B., and Cho, S. (2016).
\newblock Detecting financial misstatements with fraud intention using
  multi-class cost-sensitive learning.
\newblock {\em Expert Systems with Applications}, 62:32--43.

\bibitem[Lee et~al., 2004]{lee2004multicategory}
Lee, Y., Lin, Y., and Wahba, G. (2004).
\newblock Multicategory support vector machines: Theory and application to the
  classification of microarray data and satellite radiance data.
\newblock {\em Journal of the American Statistical Association},
  99(465):67--81.

\bibitem[Liu et~al., 2017]{liu2017hybrid}
Liu, Z., Tang, D., Cai, Y., Wang, R., and Chen, F. (2017).
\newblock A hybrid method based on ensemble welm for handling multi class
  imbalance in cancer microarray data.
\newblock {\em Neurocomputing}, 266:641--650.

\bibitem[Mahmudah et~al., 2021]{mahmudah2021machine}
Mahmudah, K.~R., Purnama, B., Indriani, F., and Satou, K. (2021).
\newblock Machine learning algorithms for predicting chronic obstructive
  pulmonary disease from gene expression data with class imbalance.
\newblock In {\em BIOINFORMATICS}, pages 148--153.

\bibitem[Mohammad, 2020]{mohammad2020improved}
Mohammad, R. M.~A. (2020).
\newblock An improved multi-class classification algorithm based on association
  classification approach and its application to spam emails.
\newblock {\em IAENG International Journal of Computer Science}, 47(2).

\bibitem[M\"{u}hlenbein, 1997]{muhlenbein1997GA}
M\"{u}hlenbein, H. (1997).
\newblock Genetic algorithms.
\newblock In Aarts, E.~H. and Lenstra, J.~K., editors, {\em Local Search in
  Combinatorial Optimization}, pages 137--172. Princeton University Press.

\bibitem[Pazzani et~al., 1994]{pazzani1994reduce}
Pazzani, M., Merz, C., Murphy, P., Ali, K., Hume, T., and Brunk, C. (1994).
\newblock Reducing misclassification costs.
\newblock In {\em ICML 1994: Proceedings of the Eleventh International
  Conference on Machine Learning}, pages 217--225. Morgan Kaufman Publishers
  Inc.: San Francisco, CA.

\bibitem[Pedregosa et~al., 2011]{pedregosa2011scikit}
Pedregosa, F., Varoquaux, G., Gramfort, A., Michel, V., Thirion, B., Grisel,
  O., Blondel, M., Prettenhofer, P., Weiss, R., Dubourg, V., Vanderplas, J.,
  Passos, A., Cournapeau, D., Brucher, M., Perrot, M., and Duchesnay, E.
  (2011).
\newblock Scikit-learn: machine learning in {P}ython.
\newblock {\em Journal of Machine Learning Research}, 12:2825--2830.

\bibitem[Schapire and Singer, 1999]{schapire1999improv}
Schapire, R.~E. and Singer, Y. (1999).
\newblock Using boosting algorithms using confidence-rated predictions.
\newblock {\em Machine Learning}, 37:297--336.

\bibitem[Snoek et~al., 2012]{snoek2012practical}
Snoek, J., Larochelle, H., and Adams, R.~P. (2012).
\newblock Practical {B}ayesian optimization of machine learning algorithms.
\newblock In {\em Advances in Neural Information Processing Systems}, pages
  2951--2959, New York, USA. Curan Associates Inc.

\bibitem[So et~al., 2021]{so2021cost}
So, B., Boucher, J.-P., and Valdez, E.~A. (2021).
\newblock Cost-sensitive multi-class adaboost for understanding driving
  behavior based on telematics.
\newblock {\em ASTIN Bulletin: The Journal of the IAA}, 51(3):719--751.

\bibitem[Sun et~al., 2007]{sun2007cost}
Sun, Y., Kamel, M.~S., Wong, A.~K., and Wang, Y. (2007).
\newblock Cost-sensitive boosting for classification of imbalanced data.
\newblock {\em Pattern Recognition}, 40(12):3358--3378.

\bibitem[Talpur and O’Sullivan, 2020]{talpur2020multi}
Talpur, B.~A. and O’Sullivan, D. (2020).
\newblock Multi-class imbalance in text classification: A feature engineering
  approach to detect cyberbullying in twitter.
\newblock In {\em Informatics}, volume~7, page~52. Multidisciplinary Digital
  Publishing Institute.

\bibitem[Tanha et~al., 2020]{tanha2020boosting}
Tanha, J., Abdi, Y., Samadi, N., Razzaghi, N., and Asadpour, M. (2020).
\newblock Boosting methods for multi‑class imbalanced data classifcation: an
  experimental review.
\newblock {\em Journal of Big Data}, 7:1--47.

\bibitem[Yuan et~al., 2018]{yuan2018regularized}
Yuan, X., Xie, L., and Abouelenien, M. (2018).
\newblock A regularized ensemble framework of deep learning for cancer
  detection from multi-class, imbalanced training data.
\newblock {\em Pattern Recognition}, 77:160--172.

\bibitem[Zhu et~al., 2009]{zhu2009mclass}
Zhu, J., Zou, H., Rossett, S., and Hastie, T. (2009).
\newblock Multi-class {A}da{B}oost.
\newblock {\em Statistics and Its Interface}, 2:349--360.

\end{thebibliography}

\end{document}